\documentclass{article}

\usepackage[margin=1.15in]{geometry}
\usepackage[utf8]{inputenc}
\usepackage{url}
\usepackage{booktabs}
\usepackage{amsfonts}
\usepackage{nicefrac}
\usepackage{microtype}
\usepackage{amsmath,amsthm,amssymb}
\usepackage{enumitem}
\usepackage{comment}
\usepackage{float}
\usepackage[colorlinks=true, allcolors=blue]{hyperref}
\usepackage{graphicx}
\usepackage{wrapfig}
\usepackage{subcaption}
\usepackage[font=small, belowskip=-.05in]{caption}
\usepackage{authblk}

\newcommand{\R}{\mathbb{R}}

\newcommand{\n}[1]{\left\| #1 \right\|}		
\newcommand{\dd}{\,\textrm d}					
\DeclareMathOperator*{\argmax}{argmax}			
\DeclareMathOperator*{\spn}{span}				
\DeclareMathOperator*{\proj}{proj}				

\begin{document}


\title{Greedy Shallow Networks: An Approach for\\Constructing and Training Neural Networks}

\author[1]{Anton Dereventsov\thanks{Corresponding author: \texttt{dereventsov@gmail.com}}}
\author[1]{Armenak Petrosyan}
\author[1,2]{Clayton~G. Webster}
\affil[ ]{}
\date{}
\maketitle
\footnotetext[1]{Computational and Applied Mathematics Group, Oak Ridge National Laboratory, Oak Ridge, TN, USA}
\footnotetext[2]{Department of Mathematics, University of Tennessee at Knoxville, Knoxville, TN, USA}
\vspace{-.5in}

\maketitle


\begin{abstract}
\noindent\emph{
We present a greedy-based approach to construct an efficient single hidden layer neural network with the ReLU activation that approximates a target function.
In our approach we obtain a shallow network by utilizing a greedy algorithm with the prescribed dictionary provided by the available training data and a set of possible inner weights.
To facilitate the greedy selection process we employ an integral representation of the network, based on the ridgelet transform, that significantly reduces the cardinality of the dictionary and hence promotes feasibility of the greedy selection.
Our approach allows for the construction of efficient architectures which can be treated either as improved initializations to be used in place of random-based alternatives, or as fully-trained networks in certain cases, thus potentially nullifying the need for backpropagation training.
Numerical experiments demonstrate the tenability of the proposed concept and its advantages compared to the conventional techniques for selecting architectures and initializations for neural networks.
}

\medskip

\noindent\textbf{Keywords:} Greedy algorithm, Integral representation, Shallow neural network, Architecture search, Ridgelet transform.

\medskip

\noindent\textbf{2000 Mathematics Subject Classification:} 68T07, 41A46, 65D15.

\noindent\textbf{2012 Computing Classification System:} 
\footnotesize
\begin{itemize}
    \item Mathematics of computing $\rightarrow$ Mathematical analysis $\rightarrow$ Functional analysis $\rightarrow$ Approximation
    \item Mathematics of computing $\rightarrow$ Mathematical analysis $\rightarrow$ Mathematical optimization $\rightarrow$ Continuous optimization $\rightarrow$ Nonconvex optimization
    \item Networks $\rightarrow$ Network architectures $\rightarrow$ Network design principles $\rightarrow$ Layering
\end{itemize}

\end{abstract}


\section{Introduction}
This paper deals with the problem of designing and training a shallow neural network for a given set of training data.
Such a problem comes up naturally in applications where neural networks are deployed and is the topic of active research, see, e.g., \cite{Reshniak:2021kt,Xie:2020dx,Xie:2019eg} and the references therein.
For an overview of the area of neural network approximation we refer the reader to the books~\cite{goodfellow2016deep,kelleher2019deep}.

Shallow neural networks are feed-forward neural networks with a single hidden layer.
Such networks are known to be universal approximators, in the sense that any continuous target function can be approximated by a shallow network with any accuracy, see e.g.~\cite{funahashi1989approximate,cybenko1989approximation}.
However, finding an appropriate (or good) network architecture for approximating a given function 
is not always a straightforward task, and theoretical estimates rely on nonlinear approximation theory \cite{DBLP:0025268} and only exists in simple settings \cite{Mhaskar2020funct,daubechies2019nonlinear,Petrosyan:2020to}.

Generally one first has to select an appropriate architecture, which is commonly done according to a heuristic rather than the given data, see e.g.~\cite{gomez2009neural,hunter2012selection}.
Even though such an approach works, it often results in unnecessarily large networks where some nodes do not actually contribute to the approximation.
\\
Next, a weight initialization scheme has to be specified.
The conventional choices, e.g.~\cite{glorot2010understanding,he2015delving,mishkin2015all} are based on sampling weights from an appropriate distribution to normalize the gradients during the backpropagation optimization.
While such approaches are relevant for deep neural networks, they do not address the issues of shallow networks.
As a result, randomly initialized neural networks often fail to achieve good approximation in practice, as explained in e.g.,~\cite{soltanolkotabi2018theoretical,hanin2019deep}.
\\
Finally, the optimization algorithm and such hyperparameters as the learning and decay rates, batch size, etc. have to be selected, see, e.g., \cite{dereventsov2020adaptive}.
This choice is generally either based on a heuristic or is dictated by an extensive manual hyperparameter tuning, see e.g.~\cite{bergstra2012random}.

Unfortunately, conventional approaches often lead the network to a bad local minimum cursed by the inherent non-convexity of the optimization problem, resulting in a low quality approximation, which is not stable and may converge slowly.
In this paper we propose a greedy selection based method for constructing shallow network approximations that do not suffer the aforementioned drawbacks and offers a trade-off between the network complexity and approximation accuracy.
We call networks obtained by this approach Greedy Shallow Networks (GSN).

The main contributions of our paper are the following:
\begin{itemize}
    \item We present an alternative method for architecture selection and the initialization/training of shallow neural networks;
    \item We develop a novel approach for reducing the size of the dictionary by utilizing the ridgelet transform and the integral representation of neural networks;
    \item We derive the procedure for selecting the network architecture based on the available training data, rather than on an established heuristic.
\end{itemize}

\subsection{Sample problem}
We demonstrate our approach in the following simple setting.
Let $f(x)$ be a one-dimensional real-valued target function we want to approximate with respect to a norm $\n{\cdot}$ by a neural network $g(x)$ with a single hidden layer of $N$ nodes, and an activation function $\sigma$, namely:
\begin{equation}\label{eq:shallow_net}
	g(x) = \sum_{n=1}^N c_n \, \sigma(a_n x + b_n),
\end{equation}
where $a_n,b_n,c_n\in\R$.
From the computational perspective it is beneficial to keep the number of nodes $N$ sufficiently small while still providing a good approximation to the target function $f(x)$.
Hence one is required to select appropriate values for $N$ and inner and outer weights $\{(a_n,b_n,c_n)\}_{n=1}^N$.
This problem can first be simplified to only selecting the number of nodes $N$ and the corresponding inner weights $\{(a_n,b_n)\}_{n=1}^N$.
Indeed, denote by $\{x_{tr},f_{tr}\} \in \mathbb{R}^{N_{tr}}$ the vectors of available training points and values respectively, then the optimal values of the outer weights $c_1,\ldots,c_N$ can be obtained easily by solving the convex optimization problem
\begin{equation}\label{eq:convex_opt}
	\min_{c_1,\ldots,c_N \in \mathbb{R}} \n{f(x_{tr}) - \sum_{n=1}^N c_n \, \sigma(a_n x_{tr} + b_n)}.
\end{equation}

In turn, we propose that the problem of selecting pairs $\{(a_n,b_n)\}_{n=1}^N$ with small $N$ can be attacked with an iterative greedy strategy, for instance, such as renown Matching Pursuit (\cite{mallat1993matching}) or Orthogonal Matching Pursuit (\cite{pati1993orthogonal}).
Indeed, the classical greedy approximation task is as follows: for an element $f$, find a sparse linear approximation by the elements of a given set $\mathcal{D}$.
The problem of constructing a network thus can be presented as follows: take sufficiently large number $M$ of possible values of the inner weights $\{(\tilde a_n, \tilde b_n)\}_{n=1}^M \subset \mathbb{R}^2$, and construct the dictionary 
\begin{equation}\label{eq:dictionary_cartesian}
	\mathcal{D} = \{\sigma(\tilde a_n x_{tr} + \tilde b_n)\}_{n=1}^M,
\end{equation}
which is then used to find an appropriate $m$-term combination $f(x) \approx \sum_{j=1}^m \gamma_j \, \sigma(\tilde a_{n_j} x + \tilde b_{n_j})$.
However, in our context the problem of constructing such a linear combination is still hardly tangible as the domain of weights $a$ and $b$ is the whole real plane $\mathbb{R}^2$, which results in an unreasonably large dictionary $\mathcal{D}$.
One way to make this approach tangible is to impose restrictions on the activation function.
Specifically, in this paper we consider the ReLU activation $\sigma(z) = \max\{z,0\}$, which appears to be the standard choice of activation in most modern network architectures.
Taking into account positive homogeneity of ReLU, we rewrite the network~\eqref{eq:shallow_net} by scaling the inner weights $\{(a_n,b_n)\}_{n=1}^N$ to be on the unit circle, i.e. $a,b \in \mathbb{R} \longrightarrow \bar a, \bar b \in \mathbb{S}^1 \subset \mathbb{R}^2$ such that
\[
	g(x) = \sum_{n=1}^N w_n \, \sigma(\bar a_n x + \bar b_n)
\]
with $w_n = c_n / \sqrt{a_n^2 + b_n^2}$.
The advantage of the above formulation is that each node is described by the single point $(\bar a_n, \bar b_n)$ on the circle $\mathbb{S}^1 \subset \mathbb{R}^2$ rather than a point on the plane $\mathbb{R}^2$.
Hence the dictionary~\eqref{eq:dictionary_cartesian} can now be obtained by only sampling the points from the circle rather than the whole plane.
Next, we employ a greedy algorithm with the constructed dictionary $\mathcal{D}$ to find an $N$-term approximation to the target function $f$.
Finally, once an appropriate approximation $f(x) \approx \sum_{n=1}^N \gamma_n \, \sigma(\bar a_n x + \bar b_n)$ is constructed, the optimal output weights $c_1,\ldots,c_N$ can be found via the convex minimization problem~\eqref{eq:convex_opt}.

\subsection{Proposed approach}\label{sec:gsn}
In this section we summarize the proposed approach for a case of general multivariate target function $f : \mathbb{R}^d \to \mathbb{R}$.
Denote by $f_{tr} \in \mathbb{R}^{N_{tr}}$ the vector of provided training values.
Then the process of constructing the GSN consists of the following steps:
\begin{enumerate}[label=\text{GSN }\arabic*., leftmargin=.5in]
	\item Sample points $\{(\bar a_j, \bar b_j)\}_{j\in\Lambda}$ from the sphere $\mathbb{S}^d$ and construct the dictionary
		\[
			\mathcal{D} = \{\sigma(\bar a_j x_{tr} + \bar b_j)\}_{j\in\Lambda} \subset \mathbb{R}^{N_{tr}};
		\]
	\item Optionally, reduce the dictionary size, e.g. via the method proposed in Section~\ref{sec:dictionary_reduction};
	\item Use the Orthogonal Greedy Algorithm (defined in Section~\ref{sec:oga}) to select the number of nodes $N$ and the corresponding inner weights $(\bar a_n, \bar b_n)$, $1 \le n \le N$;
	\item Compute the outer weights $c_n$ by solving the convex optimization problem on the training data $\{x_{tr},f_{tr}\}$
		\[
			\min_{c_1,\ldots,c_N \in \mathbb{R}} \n{f_{tr} - \sum_{n=1}^N c_n \, \sigma(\bar a_n x_{tr} + \bar b_n)};
		\]
	\item Optionally, further train the network via a backpropagation algorithm using the constructed weights $\{(c_n, \bar a_n, \bar b_n)\}_{n=1}^N$ as an initialization.
\end{enumerate}
For convenience of the presentation in this paper we consider one-dimensional output functions $f : \mathbb{R}^d \to \mathbb{R}$; however our method is also applicable in a case of multi-dimensional output target function $f : \mathbb{R}^d \to \mathbb{R}^t$ with trivial modifications.
Most notably, the convex optimization problem in the step GSN~4 would have to be solved for $c_1,\ldots,c_N \in \mathbb{R}^t$.

\subsection{Related work}
Although greedy selection methods for neural network approximations has been studied in the past by various authors, e.g.~\cite{donahue1997rates,barron2008approximation}, they were primarily used as a theoretical tool rather than the practical approach for constructing neural networks.

Another approach that proposes the discretization of the inner weights space is the Random Vector Functional-Link (or RVFL) network, e.g.~\cite{pao1994learning,pao1995functional}, where the inner weights are generated randomly and the outer weights are found via a least square regularization, however such approach results in a large number of nodes and thus an unnecessarily complex networks.

The dictionary reduction step in our approach (defined in Section~\ref{sec:int_repr}) relies on neural network integral representations, particularly on the ridgelet transform.
Similar approach is considered in~\cite{sonoda2013nonparametric}, where the authors suggest to use the values of the ridgelet transform to derive a probability distribution from which the inner weights are randomly selected.
We replace the random selection step with a greedy selection step which, even though being more expensive as a selection method, results in a better approximation with a much fewer number of nodes.
Moreover, the integral representation is completely optional in our approach, it is used only to decrease the dictionary size to accelerate the realization of the greedy algorithm.

Alternative methods for discretizing integral representations have been considered by various authors.
In particular, in~\cite{bengio2006convex} the authors employ a greedy method to discretize the solution of a certain convex optimization problem, while the same problem is solved via the conditional gradient algorithm in~\cite{bach2017breaking}.
A related Monte--Carlo discretization method for integral representations of radial basis function (RBF) neural networks  is considered in \cite{mhaskar2004tractability}.

\section{Greedy selection}\label{sec:greedy_selection}
A problem of greedy approximation can be stated as follows: find the best $N$-term approximation for a given element $f$ of a Banach space $\mathcal{X}$ with respect to the dictionary $\mathcal{D}$ (a set of elements that is dense in $\mathcal{X}$), i.e.
\begin{gather*}
    \text{Find elements}\ \ 
    g_1, \ldots, g_N \in \mathcal{D}
    \ \ \text{and coefficients}\ \ 
    c_1, \ldots, c_N \in \mathbb{R}
    \\
    \ \ \text{such that}\ \ 
    \n{f - \sum_{n=1}^N c_n g_n}_\mathcal{X}
    = \min_{\substack{g_1, \ldots, g_N \in \mathcal{D}\\c_1, \ldots, c_N \in \mathbb{R}}} \n{f - \sum_{n=1}^N c_n g_n}_\mathcal{X}.
\end{gather*}
Greedy algorithms offer ways of approximately solving such a problem by iteratively expanding the support of the allowed linear combinations.
Despite such an abstract setting, greedy algorithms are known to be successfully applied to many practical problems with various modifications, see e.g.~\cite{tropp2007signal,barron2008approximation,binev2011convergence,foucartrauhut2013}.

In this paper we are using the Orthogonal Greedy Algorithm (OGA), which is defined below.
For the purpose of presentation we only state the algorithm in Euclidean space setting; however, like other greedy algorithms, it is applicable in a general setting of infinite-dimensional Banach spaces, thus allowing for an extension of our approach for other problem settings.
The suitability of employing greedy algorithm for selecting the nodes is supported by theoretical research (including such results as convergence rate estimates and the provable near-best sparse approximation guarantees) in the approximation theory community, which is beyond the scope of this paper.
For an in-depth overview of the field of greedy algorithms we refer the reader to the book~\cite{temlyakov2011greedy}.

\subsection{Orthogonal Greedy Algorithm (OGA)}\label{sec:oga}
Let the set $\mathcal{D}$ be a dictionary in $\mathbb{R}^{N_{tr}}$, i.e. the elements of $\mathcal{D}$ are normalized and $\overline{\spn\mathcal{D}} = \mathbb{R}^{N_{tr}}$.
Let $f \in \mathbb{R}^{N_{tr}}$ be an element to be approximated and let $\proj(f,G)$ denote the orthogonal projection of $f$ onto the subspace $G \subset \mathbb{R}^{N_{tr}}$.
Then the OGA constructs sparse approximation of $f$ with respect to $\mathcal{D}$ in the following way.

\medskip
{\bf OGA:} Set $f_0 = f$ and on each iteration $N \ge 1$ perform the following steps:
\begin{enumerate}
	\item Choose an element $g_N \in \mathcal{D}$ that maximizes the inner product with $f_{N-1}$
    	\[
            g_N = \argmax_{g \in \mathcal{D}} \big|\big\langle f_{N-1},\, g \big\rangle\big|;
        \]
    \item Compute the orthogonal projection of $f$ onto $\spn\{g_1,\ldots,g_N\}$
        \[
            \text{find}\ \ 
            c_1,\ldots,c_N \in \mathbb{R}
            \ \ \text{such that}\ \ 
            \sum_{n=1}^N c_n g_n = \proj(f, \spn\{g_1,\ldots,g_N\});
        \]
	\item Set the next remainder $f_N$ as
        \[
            f_N = f - \proj(f, \spn\{g_1,\ldots,g_N\}) = f - \sum_{n=1}^N c_n g_n.
        \]
\end{enumerate}
This procedure can be continued until either the sufficient approximation accuracy (measured by $\n{f_N}$) is reached, or the maximal number of iterations is completed, hence allowing for the trade-off between the representation sparsity and the approximation accuracy.

We note that in a more general setting the performance of greedy algorithms suffers from the dictionary size, which is typically large in high dimensions.
In particular, in the case of infinitely-dimensional spaces, step $1$ of the OGA might not even be feasible.
Conventional ways to overcome this issue would be to consider a {\it weak} version of a greedy algorithm or to perform a {\it restricted depth} dictionary search, see e.g.~\cite{temlyakov2000weak,temlyakov2005greedy}.
Additionally, the computations required in steps $1$ and $2$ can be relaxed by considering an {\it approximate} version of a greedy algorithm in which the computations are performed not precisely but with allowed inaccuracies of controllable magnitude, see e.g.~\cite{gribonval2001approximate,galatenko2005generalized}.
Such approaches are well studied and are known to simplify the realization of a greedy algorithm.

In the current manuscript we focus on the case of finitely-dimensional Hilbert spaces and consider the original formulation of the OGA with the exact calculations in steps $1$--$3$.
The main challenge in our context is the large dimensionality of the dictionary $\mathcal{D}$, which we address by deriving a novel approach for domain clustering based on the ridgelet transform.
Namely, in Section~\ref{sec:dictionary_reduction} we propose a method for reducing the dictionary size which is based on an integral representation of a neural network.

\section{Integral representations}\label{sec:int_repr}
In this section we propose a method for the dictionary size reduction, based on an integral representation of the network, that restricts the sampling domain for the inner weights from the $d$-sphere $\mathbb{S}^d$ to its subset.
Namely, for a given target function $f : \mathbb{R}^d \to \mathbb{R}$, we consider representations of the form
\begin{equation}\label{eq:dual_ridge_rep}
	f(x) = \int\limits_{\mathbb{R}^d \times \mathbb{R}} c(a,b) \, \sigma(a \cdot x+b) \dd a\dd b,
\end{equation}
called the neural network integral representation of $f$ with the kernel $c(a,b)$.
Note that for a given function $f$ there can be infinitely many suitable kernels $c$ such that \eqref{eq:dual_ridge_rep} holds.
The idea of the dictionary reduction comes from the observation that the inner weights should be sampled only from the support of the kernel $c$ rather than the whole space.

To give a concrete example and to demonstrate the dictionary reduction property, we consider here the representation based on the ridgelet transform.
In our numerical experiments we observe that the proposed technique allows for a dictionary reduction by approximately $50\%$ in case of one-dimensional target function $f$ (examples~\ref{sec:ex1} and~\ref{sec:ex2}), and by $70\%$ in case of two-dimensional one (examples~\ref{sec:ex3} and~\ref{sec:ex4}), without a noticeable affect on the quality of the constructed network.

We also note that there are more efficient ways to achieve search space reduction, such as those based on more fitting integral representations, see e.g.~\cite{ongie2019function,Petrosyan:2020to}.
However the overview of such methods is outside the scope of the current paper and will be discussed in future publications.

\subsection{Dictionary reduction procedure}\label{sec:dictionary_reduction}
One of the integral representations~\eqref{eq:dual_ridge_rep} that we consider below originates from the harmonic analysis perspective to shallow neural networks, see, e.g.~\cite{candes1999harmonic}, and employs the ridgelet transform given by
\begin{equation}\label{form:ridgelet_def}
	\mathcal{R}_\tau f(a,b) = \int\limits_{\mathbb{R}^d} f(x) \, \overline{\tau(a \cdot x + b)} \dd x,
\end{equation}
with some choice of function $ \tau : \mathbb{R} \to \mathbb{C}$.
It is known that under the admissibility condition
\begin{equation*}\label{eq:admissibility_constant}
	 (2\pi)^{d-1} \int\limits_{-\infty}^\infty \frac{\hat\sigma(z) \, \overline{\hat\tau(z)}}{|z|^d} \dd z=1,
\end{equation*}
where $\hat \sigma, \hat \tau$ are the Fourier transforms of $\sigma$ and $\tau$, for a large class of target functions $f$ (see e.g.~\cite{sonoda2017neural} for details) the following reconstruction formula holds
\begin{equation}\label{rep:ridgelet_form}
	f(x) = \int\limits_{\mathbb{R}^d \times \mathbb{R}} \mathcal{R}_\tau f(a,b) \, \sigma(a \cdot x+b) \dd a\dd b.
\end{equation} 
In particular, the ReLU activation $\sigma(z) = \max\{z,0\}$ forms an admissible pair with the function
\[
	\tau(z)
	= -\frac{(\exp(-z^2/2))^{\prime\prime\prime\prime}}{2 (2\pi)^{d-1/2}}
	= -\frac{z^4 - 6z^2 + 3}{2 (2\pi)^{d-1/2}} \, e^{-z^2/2},
\]
which is the function that we use in our numerical experiments.

By switching to spherical coordinates and using the positive homogeneity of the ReLU, we can rewrite the reconstruction formula~\eqref{rep:ridgelet_form} in the following way
\begin{align}
	\nonumber
	f(x)
	&= \int\limits_{\mathbb{S}^d} \int\limits_0^\infty \mathcal{R}_\tau(r \bar a,r\bar b) \, 
		r^{d+1} \dd r \, \sigma(\bar a \cdot x + \bar b) \dd (\bar a,\bar b)
	\\
	\label{eq:crf_int_repr}
	&= \int\limits_{\mathbb{S}^d} \mathcal{CR}_\tau f(\bar a,\bar b) \, \sigma(\bar a \cdot x + \bar b) \dd (\bar a,\bar b),
\end{align}
where $\mathcal{CR}_\tau f$ is what we call the collapsed ridgelet transform, defined as 
\begin{equation}\label{eq:collapsed_ridgelet}
	\mathcal{CR}_\tau f(\bar a,\bar b) = \int\limits_0^\infty \mathcal{R}_\tau(r \bar a,r\bar b) \, r^{d+1} \dd r .
\end{equation}
Now the reduced dictionary $\mathcal{D}^\prime$ can by obtained by keeping only those vectors $\sigma(\bar a_n x_{tr} + \bar b_n)$ in $\mathcal{D}$, whose weights $(\bar a_n, \bar b_n)$ satisfy $|\mathcal{CR}_\tau f(\bar a_n,\bar b_n)| > \epsilon$ with a problem dependent threshold $\epsilon > 0$.
The value of $\epsilon$ in this context controls the trade-off between the potential approximation accuracy and the computational complexity.
Namely, larger values of $\epsilon$ promote the reduced cardinality of the dictionary $\mathcal{D}^\prime$, and thus simplify the greedy selection step, however at a possible cost of approximation accuracy.

The rationale of why thresholding the collapsed ridgelet transform results in efficient dictionary size reduction is supported by our theoretical findings which state that $\mathcal{CR}_\tau f$ is well-localized.
For the purpose of this paper, however, we only support this claim with numerical experiments (Figures~\ref{fig:ex1_crf}, \ref{fig:ex2_crf}, \ref{fig:ex3_crf}, and~\ref{fig:ex4_crf}).

A particular importance of the integral representation~\eqref{eq:crf_int_repr} is that, under the assumption that $\mathcal{CR}_\tau f \in L_1(\mathbb{S}^d)$, it can be deduced that the vector of training values $f_{tr}$ is in the closure of the convex hull of the dictionary $\mathcal{D} = \{\sigma(\bar a \cdot x_{tr} + \bar b)\}_{(\bar a, \bar b) \in \mathbb{S}^d}$.
As follows from the greedy approximation theory~\cite{devore1996some}, in that case the GSN initialization described in Subsection~\ref{sec:gsn} approximates the training data $f_{tr}$ with the rate of $\mathcal{O}(N^{-1/2})$, where $N$ is the number of nodes (i.e. iterations of the greedy algorithm).
Note that this is the worst-case estimate, in practice we generally observe exponential convergence rates.

\section{Numerical experiments}\label{sec:numerics}
All numerical experiments are implemented in Python 3.6.
Neural networks are trained in Keras with the TensorFlow backend, and the rest of the computations are performed with the use of Numpy package.
For reproducibility, the Numpy random seed is set to $0$ for all presented numerical experiments.
The code used to obtain the presented numerical results is publicly available at~\url{https://github.com/sukiboo/sgn}.

\begin{wrapfigure}{r}{.3\linewidth}
	\centering
	\includegraphics[width=\linewidth]{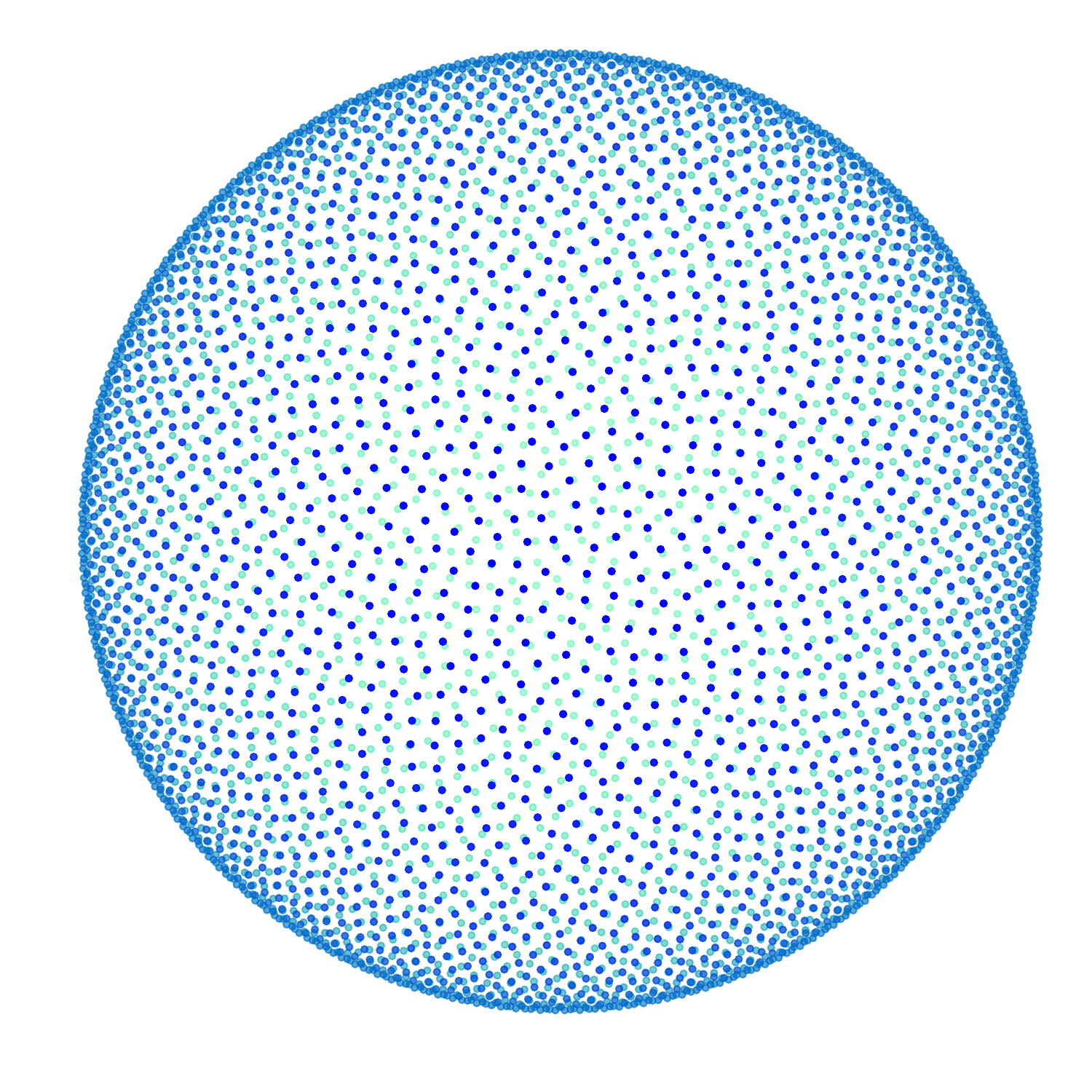}
	\caption{5,000 points on $\mathbb{S}^2$ sampled with the golden spiral method.}
	\label{fig:points_sphere}
\end{wrapfigure}

In order to construct the dictionary $\mathcal{D}$ for a $d$-dimensional target function, we sample $10,000\,d$ points from the $d$-sphere $\mathbb{S}^d$.
In case $d = 1$ we sample the points uniformly from the interval $[-\pi,\pi)$.
In case $d = 2$ we employ the golden spiral method (shown in Figure~\ref{fig:points_sphere}) to sample the points from the sphere $\mathbb{S}^2$.
For other values of $d$ we generate and normalize independent Gaussian vectors, which are uniformly distributed (in expectation) over $\mathbb{S}^d$.

For the settings $d = 1,2$ we reduce the size of the dictionary $\mathcal{D}$ via the method described in Section~\ref{sec:int_repr}.
Namely, we compute the ridgelet transform $\mathcal{R}f(r,\phi)$ by discretizing the integral~\eqref{form:ridgelet_def} on the provided training data, and the collapsed ridgelet transform $\mathcal{CR}f(\phi)$ by summing the ridgelet transform along the sampled directions $\phi \in \mathbb{S}^d$ using the formula~\eqref{eq:collapsed_ridgelet}.
The values of $\mathcal{CR}f(\phi)$ are used to reduce the dictionary $\mathcal{D}$ by discarding the sampled points $\phi$ that correspond to the values of $|\mathcal{CR}f(\phi)|$ that are smaller than the $0.1\%$ of the maximal value of $|\mathcal{CR}f(\phi)|$.

The Orthogonal Greedy Algorithm is applied to the vector of training values with respect to the dictionary $\mathcal{D}$.
The number of constructed nodes is chosen as the iteration of the greedy algorithm that minimizes the approximation error on the given validation set.
That is, the resulting network architecture is the one that minimizes the validation error as shown in Figures~\ref{fig:ex1_oga}, \ref{fig:ex2_oga}, \ref{fig:ex3_oga}, \ref{fig:ex4_oga}, and \ref{fig:ex5_oga} for the respective examples.

For each target function we construct two network approximations and compare them.
First, we obtain the GSN initialization as described in Subsection~\ref{sec:gsn}, and train it via the backpropagation.
Next, we train the neural network with random initialization with `TruncatedNormal' initial weights for both the hidden and the output layers.
Both networks are trained with the `Adam optimizer', see~\cite{kingma2014adam}, with the initial learning rate $10^{-3}$ and the exponential decay rate of $4.6 \times 10^{-4}$.
The training is performed for $10,000$ epochs to ensure that the learning is completed; plots of the training losses are presented to demonstrate that every training process is completed and an appropriate hyperparameters are selected.
Since it is known that networks with random initialization and small number of parameters are sensitive to the initial state and are prone to get stuck in a bad local minimum, in some examples we use the smaller batch sizes for randomly initialized networks than for the GSN ones; otherwise the choice of hyperparameters for the training of two networks is the same.
Each randomly initialized network we train for $10$ times and report the best result the network achieved; networks with the GSN initialization are trained once since the initialization is deterministic.

\subsection{Example 1}\label{sec:ex1}
\begin{figure}[htbp]
	\centering
	\begin{subfigure}{.32\linewidth}
		\includegraphics[width=\linewidth]{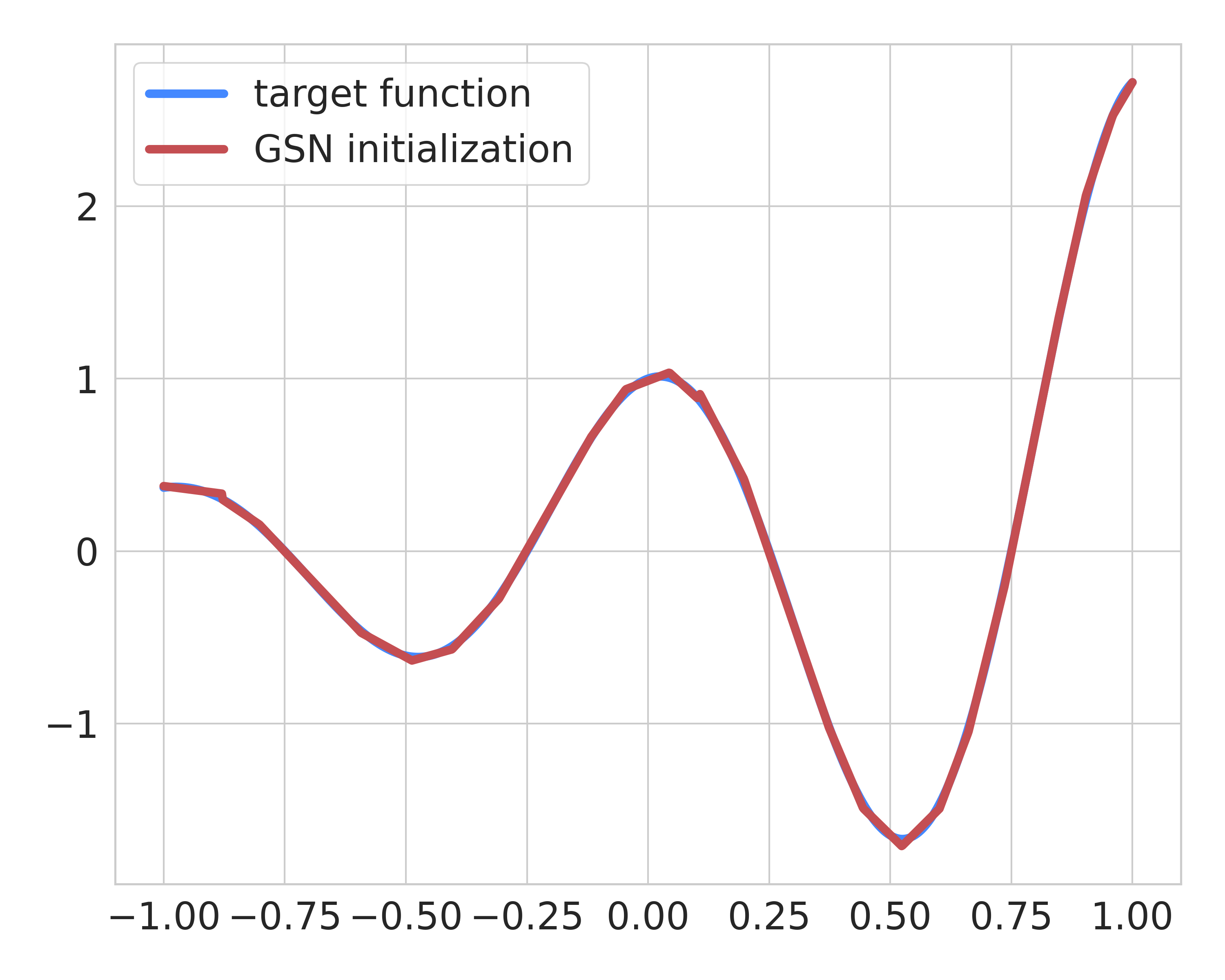}
		\caption{1.02e-02}
		\label{fig:ex1_appr_ga_init}
	\end{subfigure}
	\begin{subfigure}{.32\linewidth}
		\includegraphics[width=\linewidth]{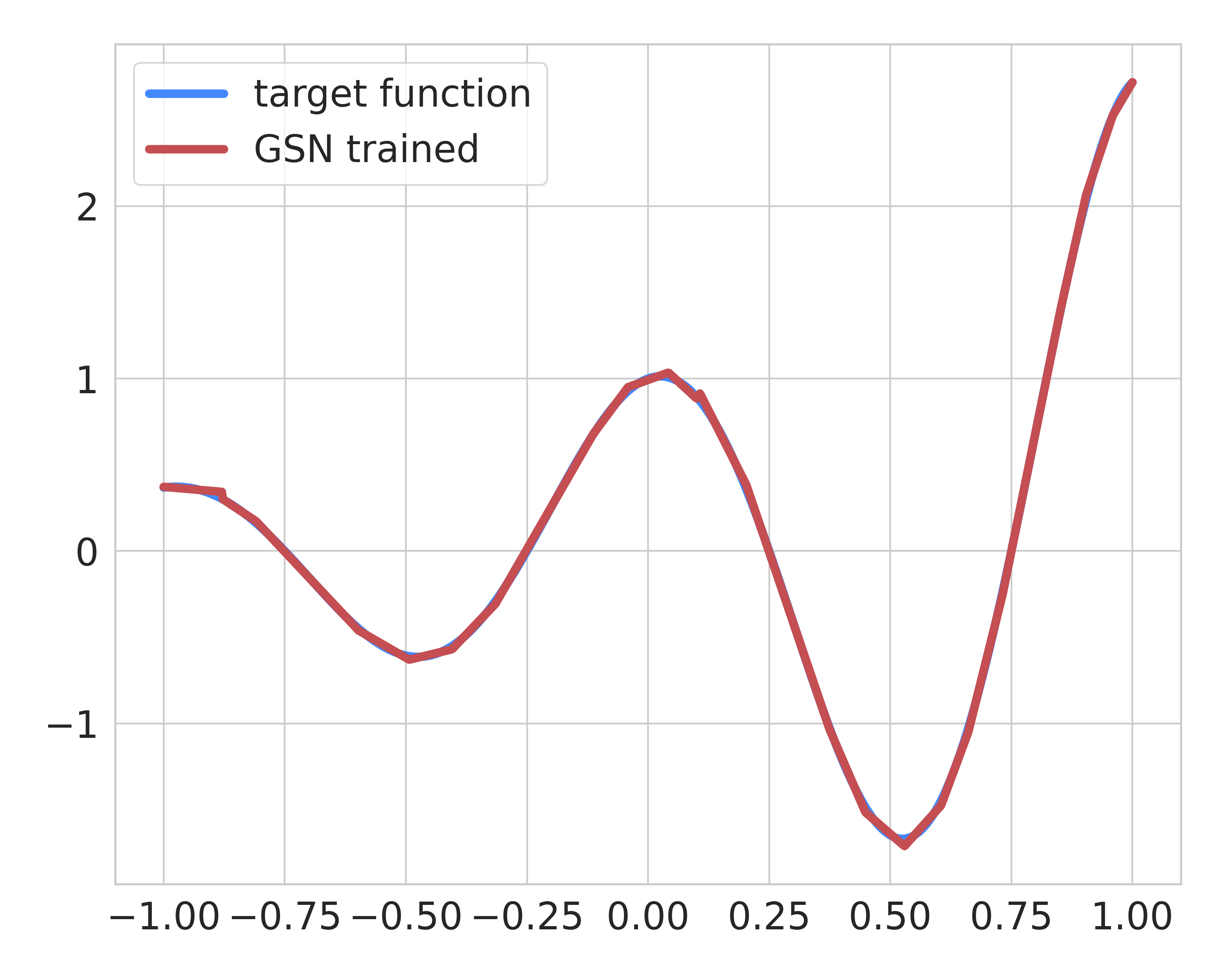}
		\caption{1.00e-02}
		\label{fig:ex1_appr_ga_train}
	\end{subfigure}
	\begin{subfigure}{.32\linewidth}
		\includegraphics[width=\linewidth]{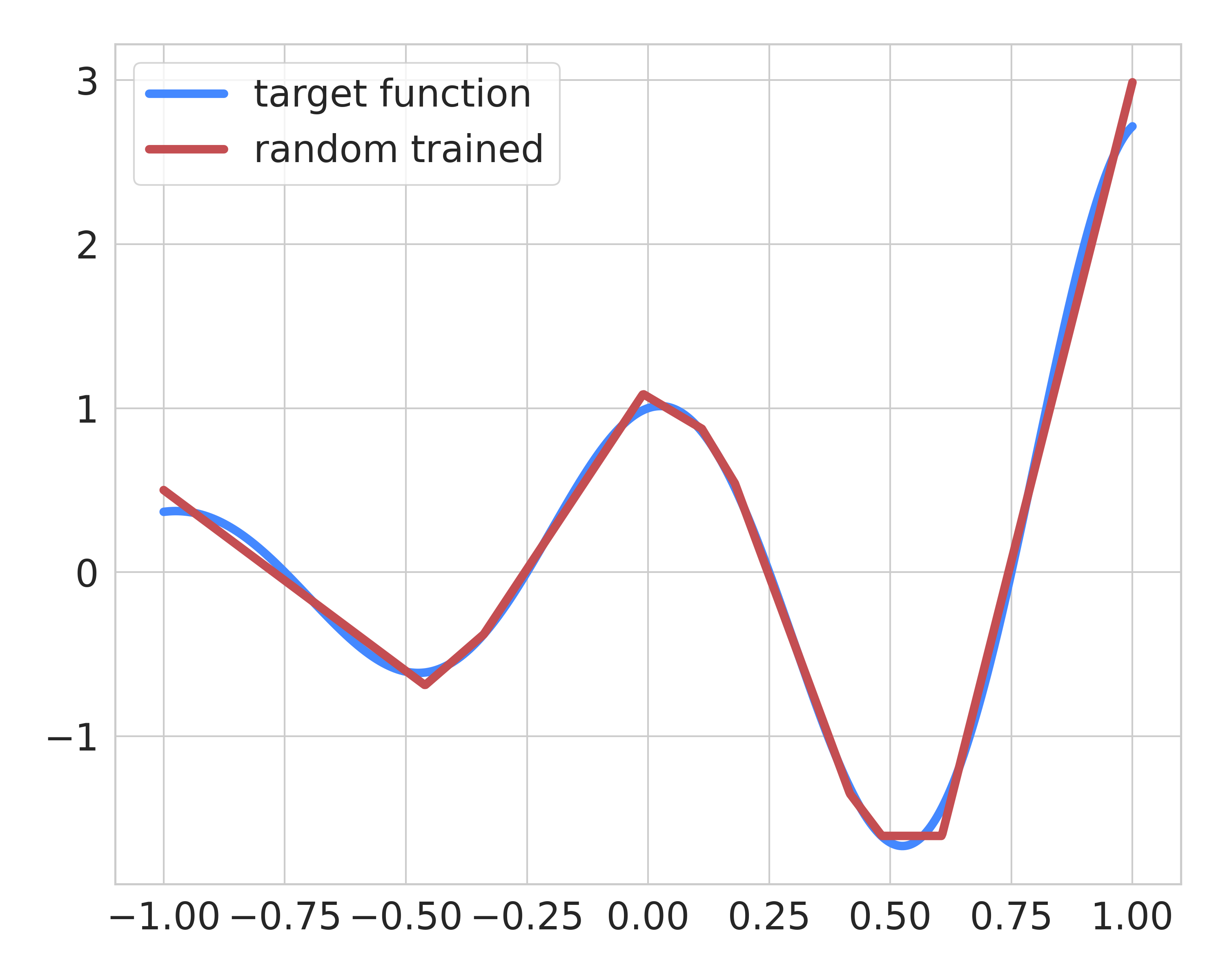}
		\caption{6.76e-02}
		\label{fig:ex1_appr_rnd}
	\end{subfigure}
	\caption{(a) GSN initialization, (b) network trained with GSN initialization, (c) network trained with random initialization. Values under the images indicate the $\ell_2$-approximation error on the test set.}
	\label{fig:ex1_appr}
\end{figure}

\begin{figure}[htbp]
	\centering
	\begin{subfigure}{.24\linewidth}
		\includegraphics[width=\linewidth,trim=3cm 2cm 3cm 3cm,clip]{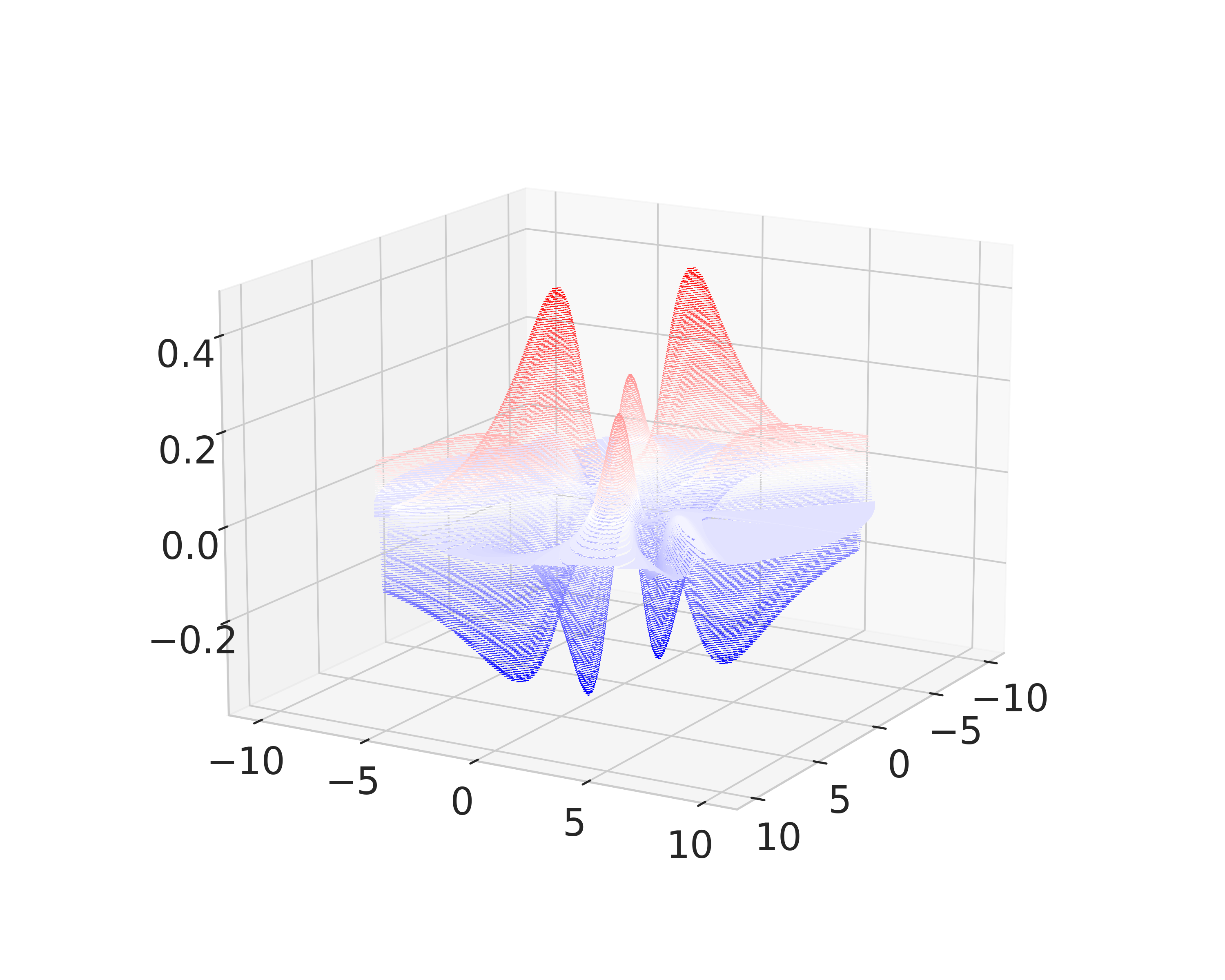}
		\caption{$\mathcal{R}f(a,b)$}
		\label{fig:ex1_rf}
	\end{subfigure}
	\begin{subfigure}{.24\linewidth}
		\includegraphics[width=\linewidth]{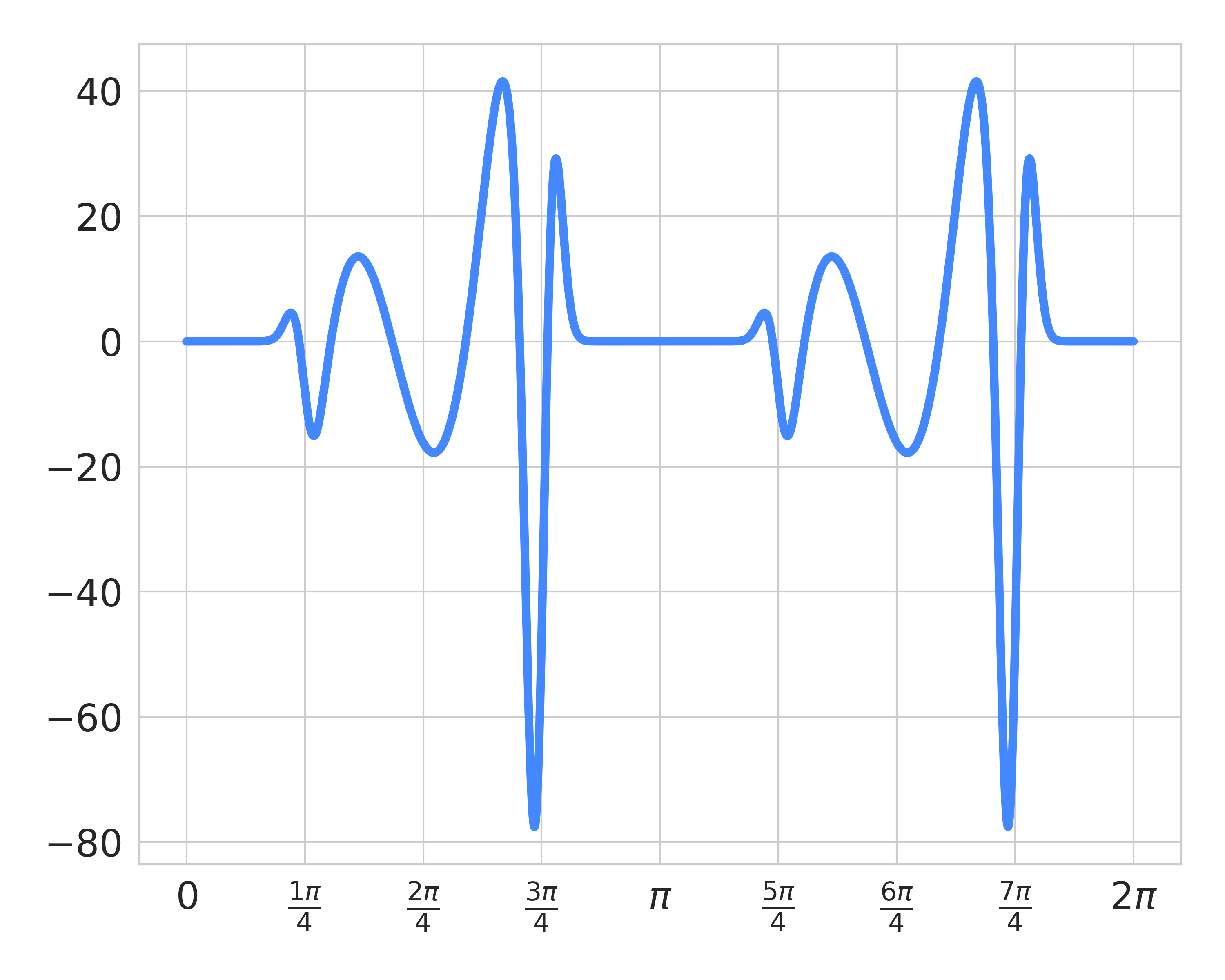}
		\caption{$\mathcal{CR}f(\phi)$}
		\label{fig:ex1_crf}
	\end{subfigure}
	\begin{subfigure}{.24\linewidth}
		\includegraphics[width=\linewidth]{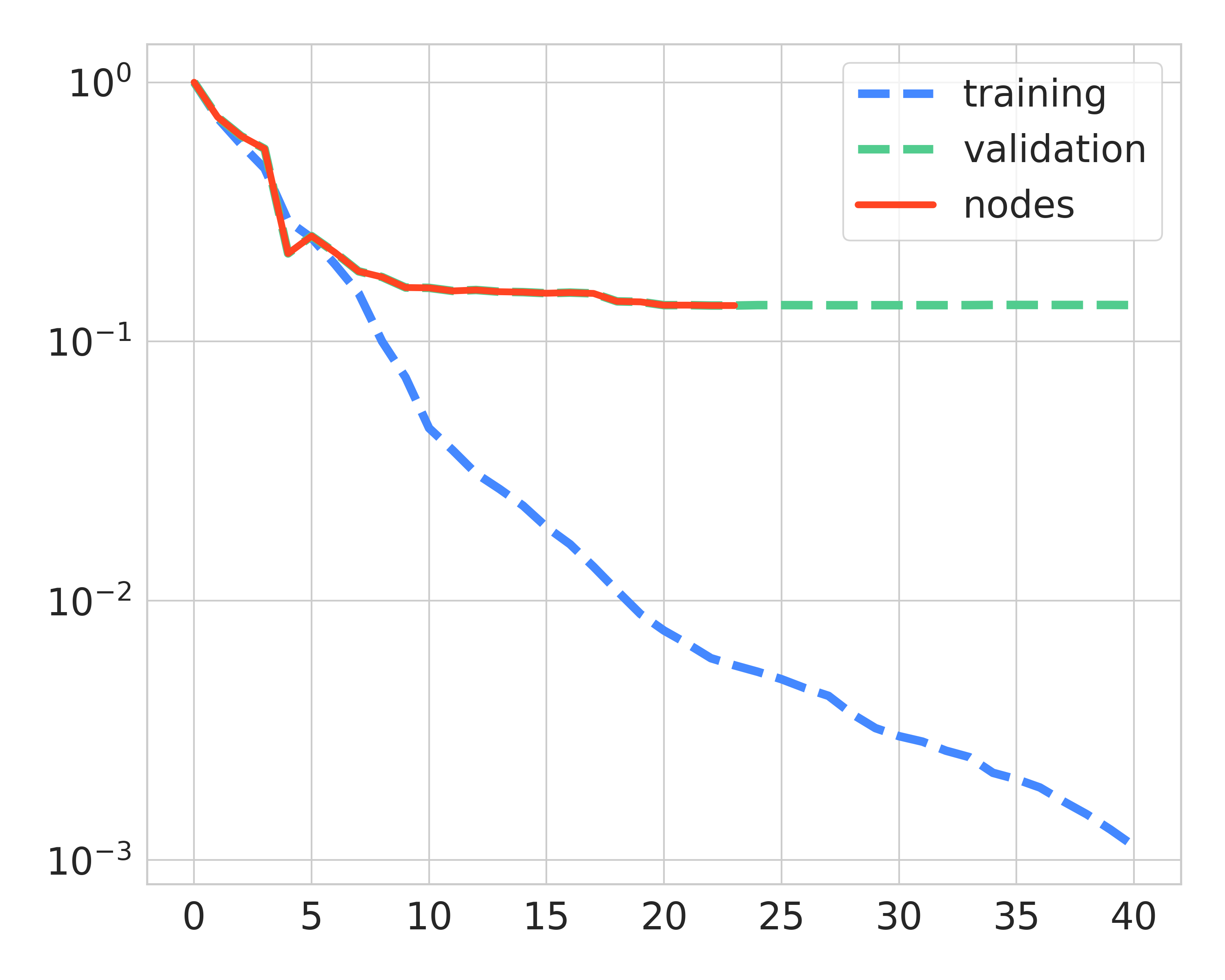}
		\caption{\#nodes: $23$}
		\label{fig:ex1_oga}
	\end{subfigure}
	\begin{subfigure}{.24\linewidth}
		\includegraphics[width=\linewidth]{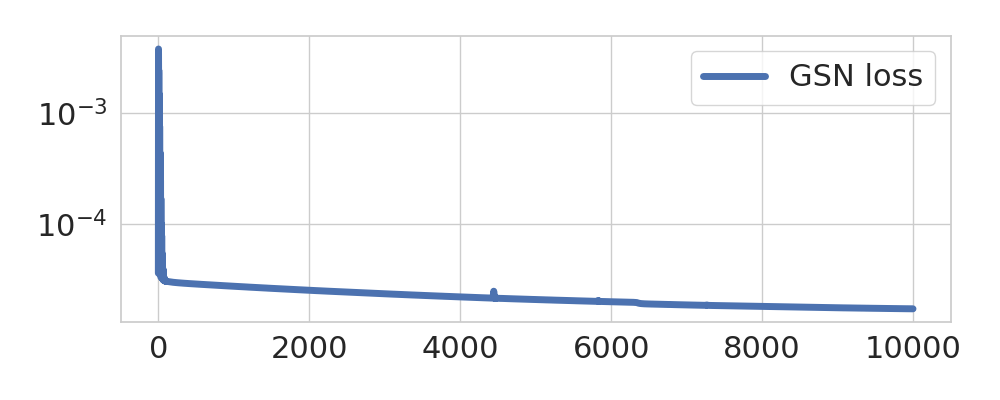}
		\includegraphics[width=\linewidth]{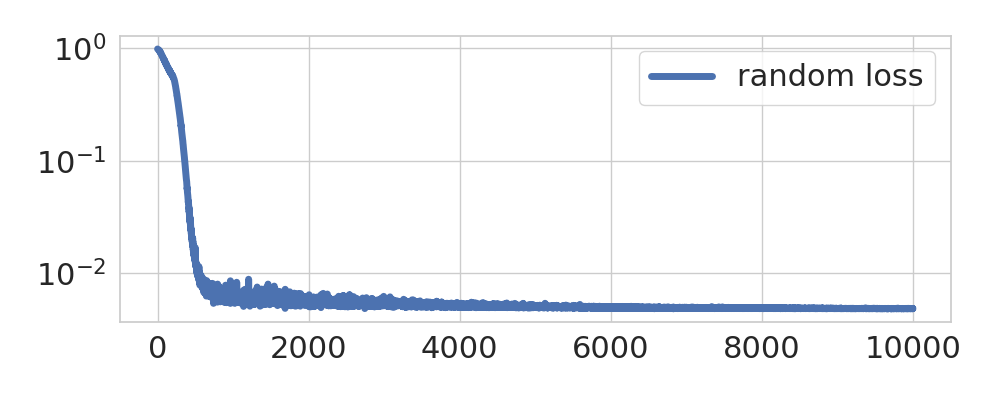}
		\caption{training losses}
		\label{fig:ex1_loss}
	\end{subfigure}
	\caption{Construction of the Shallow Greedy Network: (a) ridgelet transform of $f$, (b) collapsed ridgelet transform of $f$, (c) nodes selection by the greedy algorithm, (d) training of the neural network with GSN initialization (top) / random initialization (bottom).}
\end{figure}

Consider the target function $f : [-1,1] \to \mathbb{R}$ given by
\[
	f(x) = \cos(2\pi x) \, \exp(x).
\]
We take $50$ training points, $15$ validation points, and $1,000$ test points.
The batch size is set to be $50$ for the network with the GSN initialization and $1$ for the network with random initialization.
The resulting approximations are presented in Figure~\ref{fig:ex1_appr}.
The ridgelet and collapsed ridgelet transforms are shown in Figures~\ref{fig:ex1_rf} and~\ref{fig:ex1_crf} respectively.
In this example the number of constructed nodes is $23$ (Figure~\ref{fig:ex1_oga}).

\subsection{Example 2}\label{sec:ex2}
\begin{figure}[htbp]
	\centering
	\begin{subfigure}{.32\linewidth}
		\includegraphics[width=\linewidth]{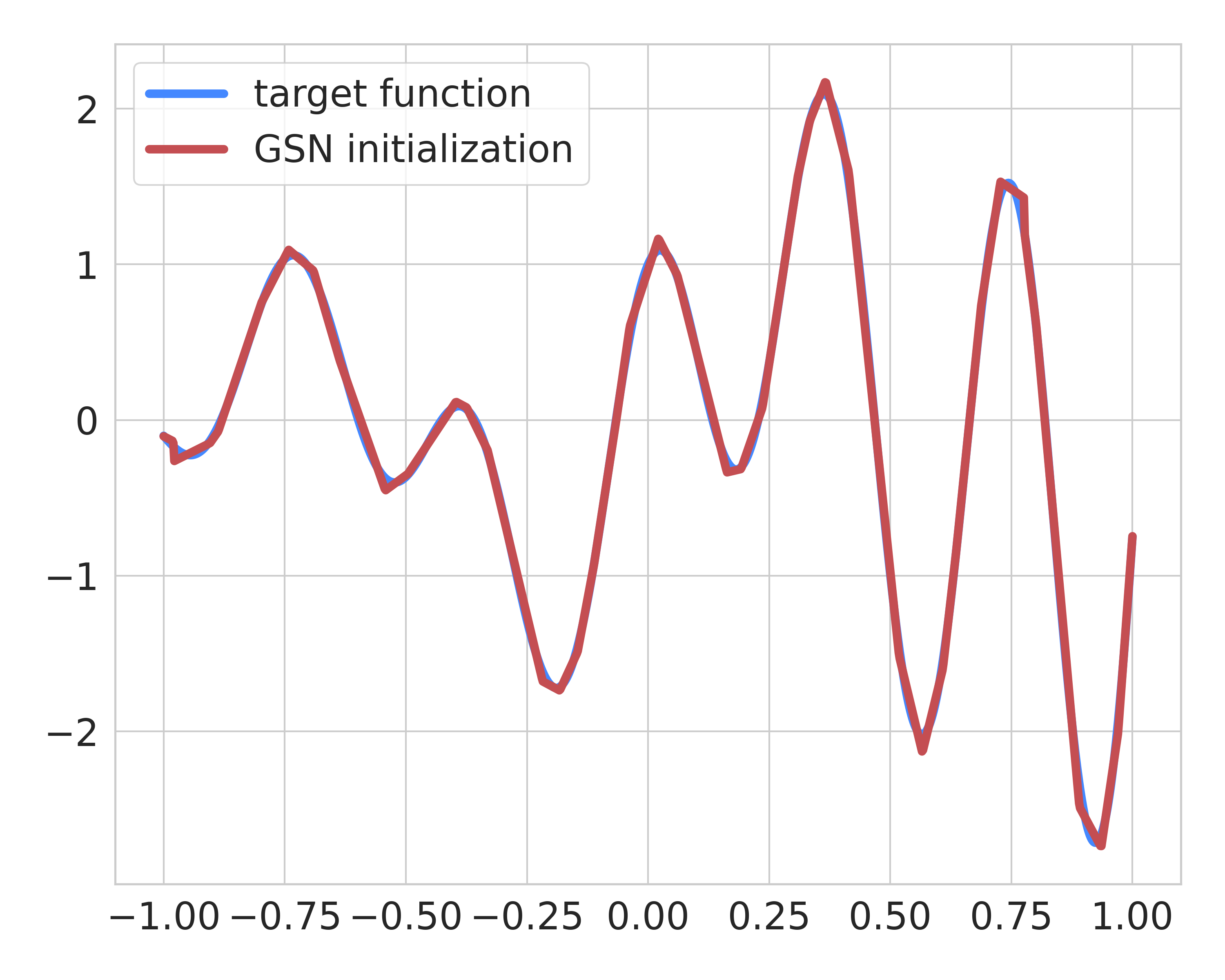}
		\caption{2.63e-02}
		\label{fig:ex2_appr_ga_init}
	\end{subfigure}
	\begin{subfigure}{.32\linewidth}
		\includegraphics[width=\linewidth]{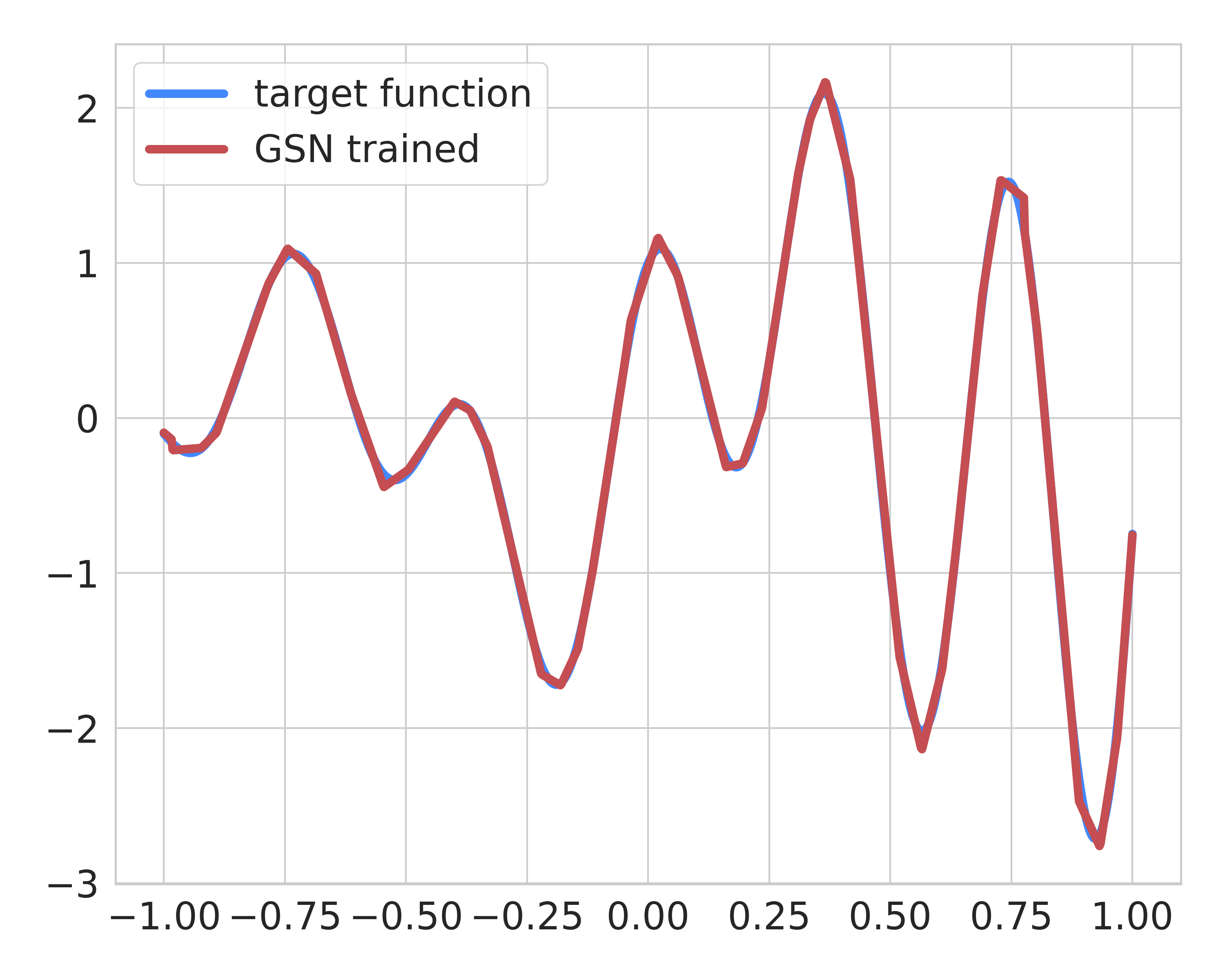}
		\caption{2.44e-02}
		\label{fig:ex2_appr_ga_train}
	\end{subfigure}
	\begin{subfigure}{.32\linewidth}
		\includegraphics[width=\linewidth]{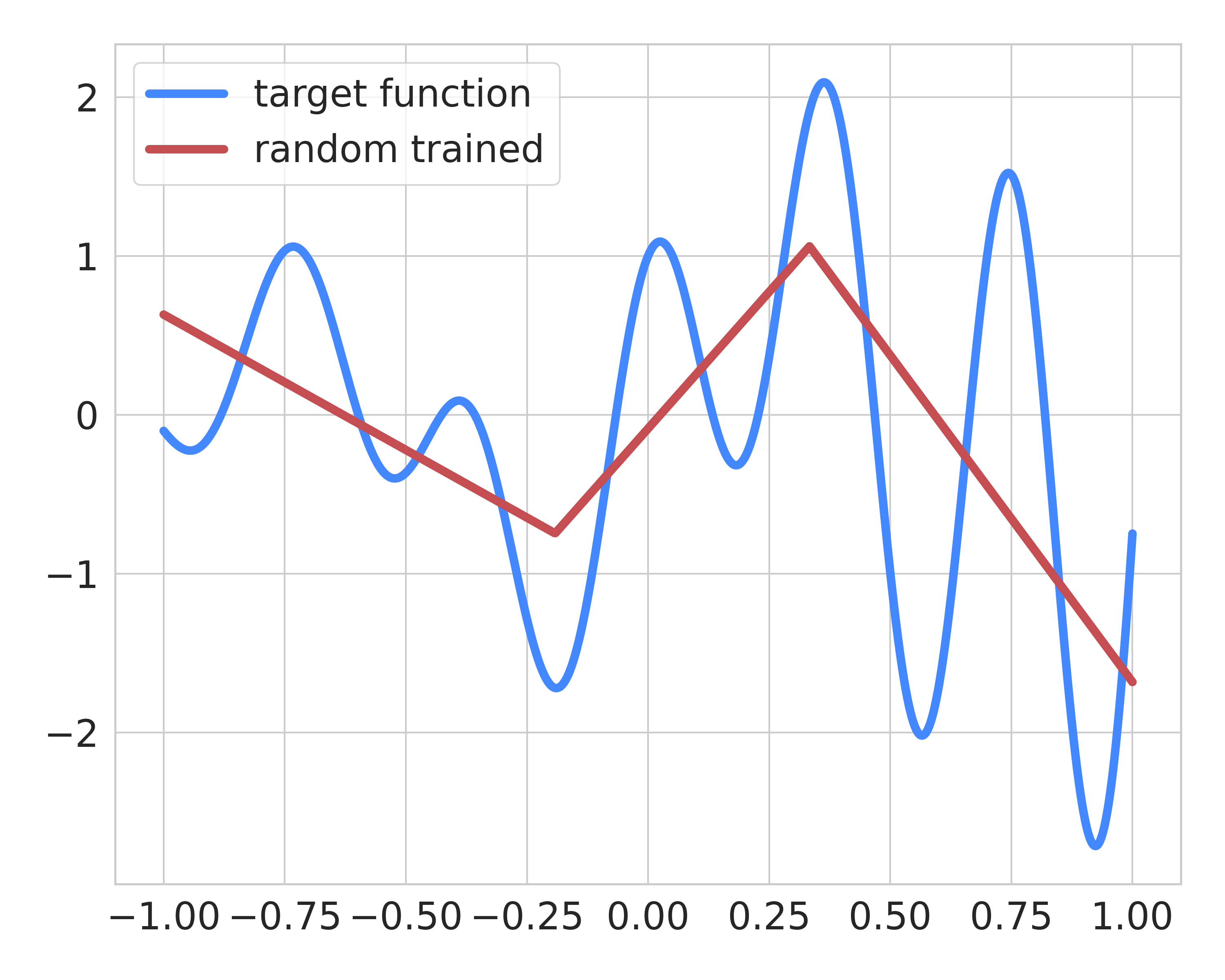}
		\caption{8.17e-01}
		\label{fig:ex2_appr_rnd}
	\end{subfigure}
	\caption{(a) GSN initialization, (b) network trained with GSN initialization, (c) network trained with random initialization. Values under the images indicate the $\ell_2$-approximation error on the test set.}
	\label{fig:ex2_appr}
\end{figure}

\begin{figure}[htbp]
	\centering
	\begin{subfigure}{.24\linewidth}
		\includegraphics[width=\linewidth,trim=3cm 2cm 3cm 3cm,clip]{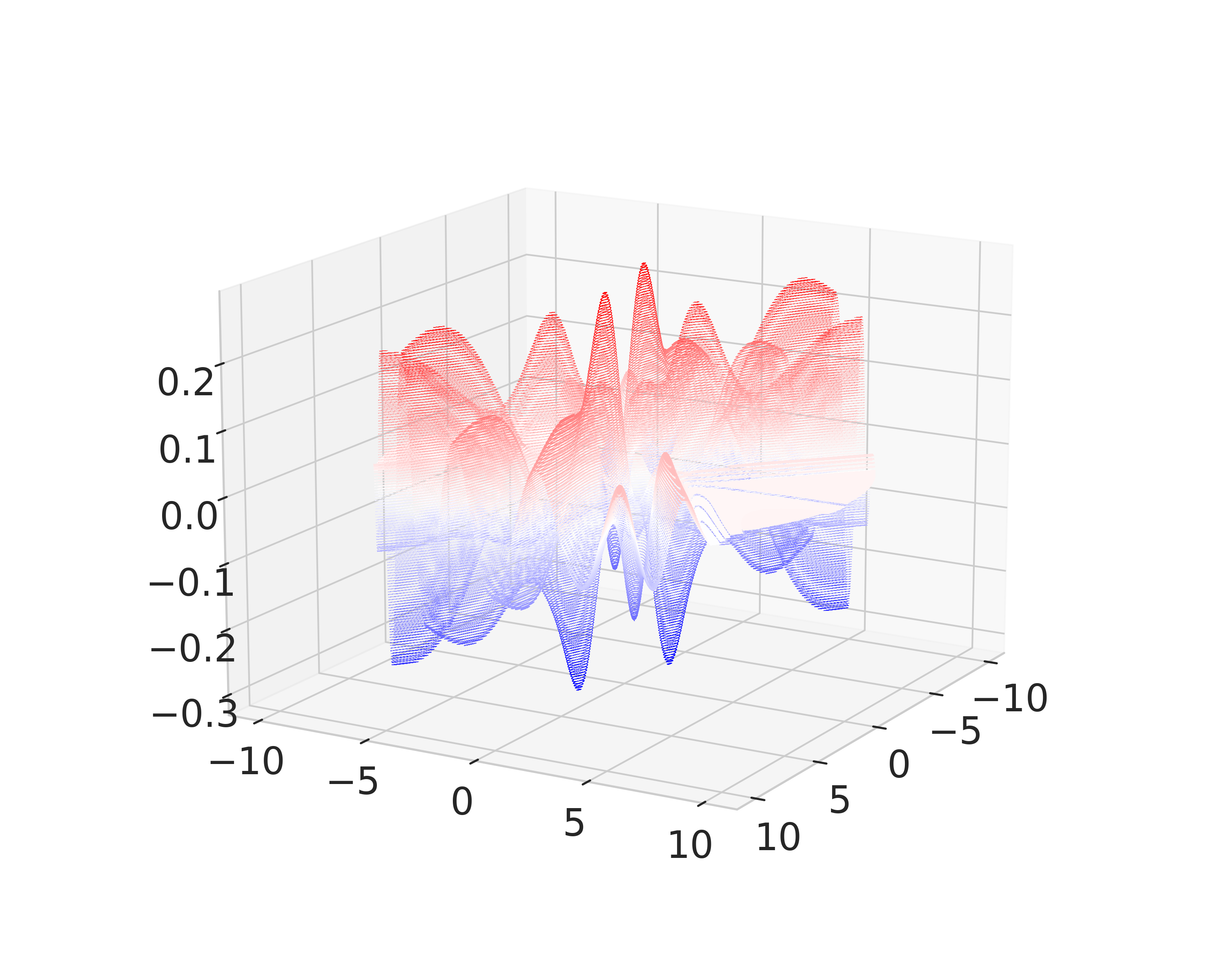}
		\caption{$\mathcal{R}f(a,b)$}
		\label{fig:ex2_rf}
	\end{subfigure}
	\begin{subfigure}{.24\linewidth}
		\includegraphics[width=\linewidth]{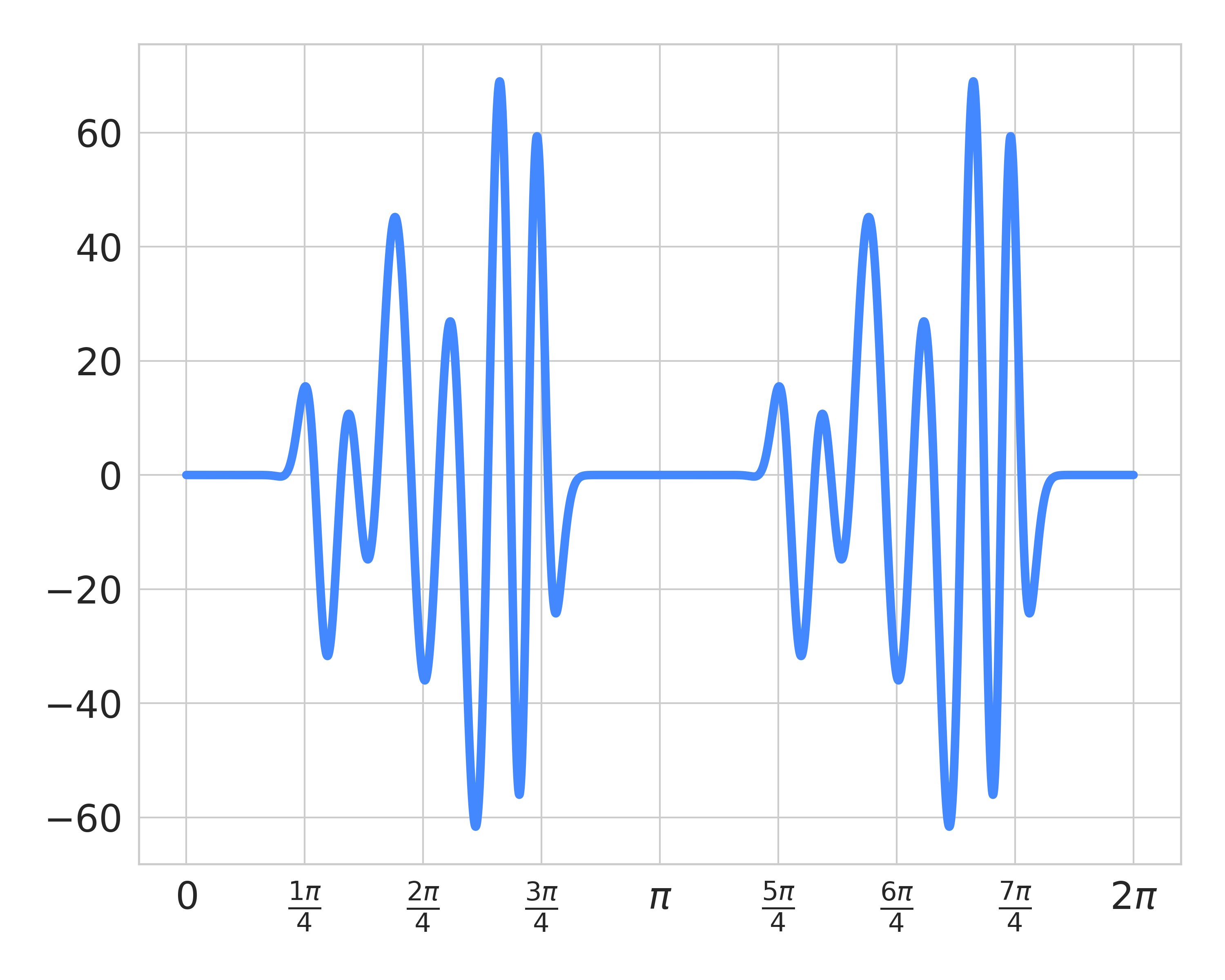}
		\caption{$\mathcal{CR}f(\phi)$}
		\label{fig:ex2_crf}
	\end{subfigure}
	\begin{subfigure}{.24\linewidth}
		\includegraphics[width=\linewidth]{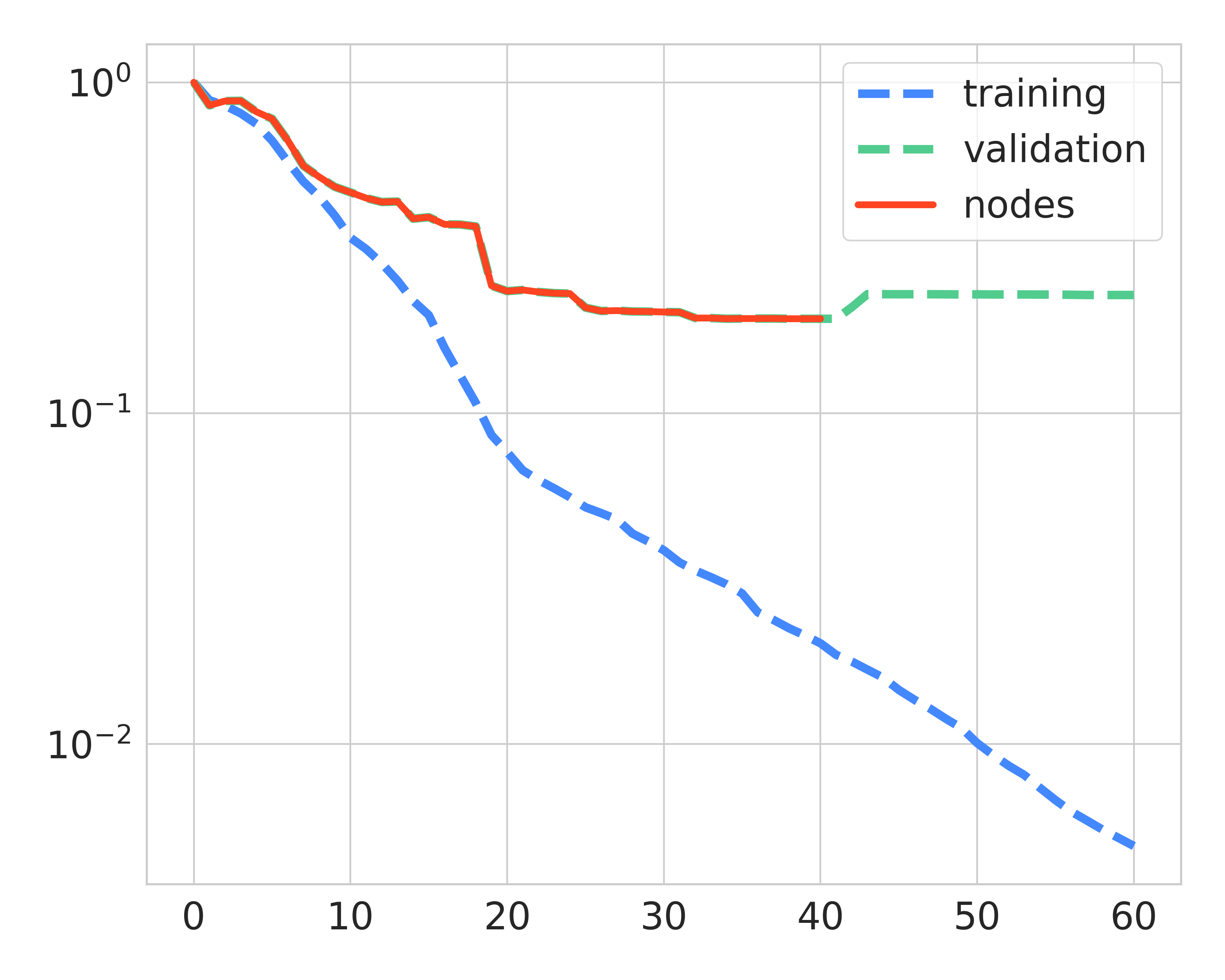}
		\caption{\#nodes: $40$}
		\label{fig:ex2_oga}
	\end{subfigure}
	\begin{subfigure}{.24\linewidth}
		\includegraphics[width=\linewidth]{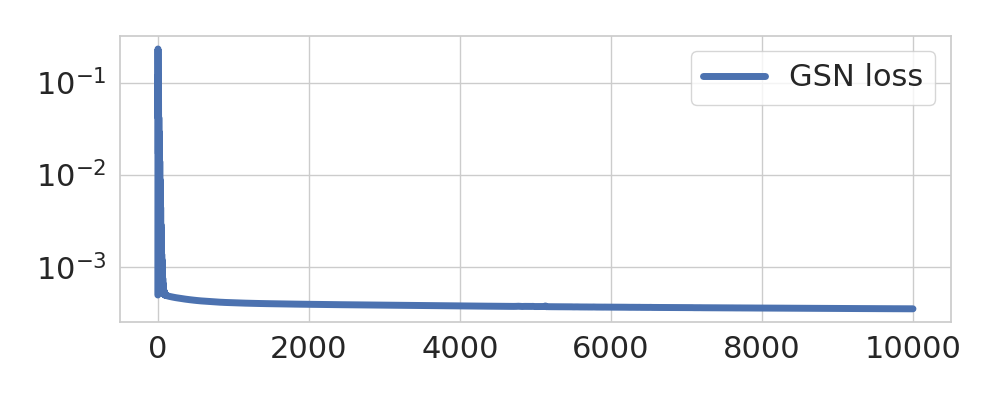}
		\includegraphics[width=\linewidth]{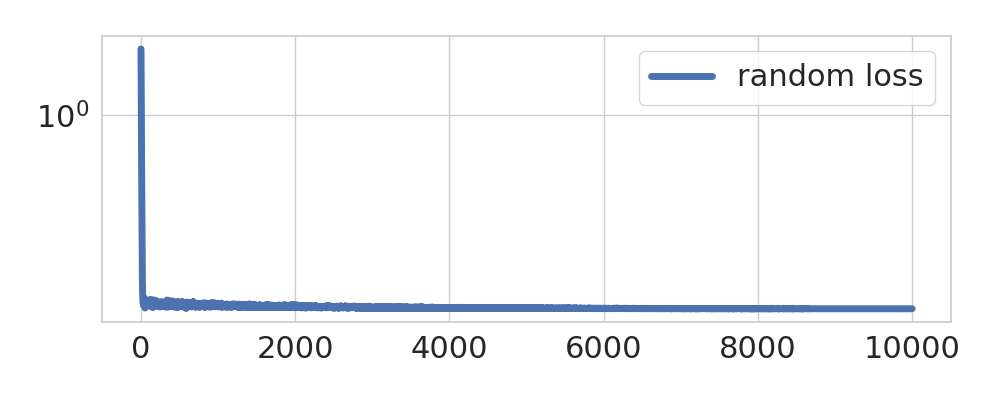}
		\caption{training losses}
		\label{fig:ex2_loss}
	\end{subfigure}
	\caption{Construction of the Shallow Greedy Network: (a) ridgelet transform of $f$, (b) collapsed ridgelet transform of $f$, (c) nodes selection by the greedy algorithm, (d) training of the neural network with GSN initialization (top) / random initialization (bottom).}
\end{figure}

Consider the target function $f : [-1,1] \to \mathbb{R}$ given by
\[
	f(x) = \sin(2\pi x) \, \exp(-x^2) + \cos(17x) \, \exp(x).
\]
We take $100$ training points, $20$ validation points, and $1,000$ test points.
The batch size is set to be $100$ for the network with the GSN initialization and $1$ for the network with random initialization.
The resulting approximations are presented in Figure~\ref{fig:ex2_appr}.
The ridgelet and collapsed ridgelet transforms are shown in Figures~\ref{fig:ex2_rf} and~\ref{fig:ex2_crf} respectively.
In this example the number of constructed nodes is $40$ (Figure~\ref{fig:ex2_oga}).
Note that in this example networks with random initialization fails to achieve a reasonable approximation despite multiple attempts.
This is likely caused by restricting the number of nodes and absence of additional regularization as explained in~\cite{hanin2019deep}.
The network with the GSN initialization, on the other hand, under the same conditions recovers every oscillation of the target function.

\subsection{Example 3}\label{sec:ex3}
\begin{figure}[htbp]
	\centering
	\begin{subfigure}{.32\linewidth}
		\includegraphics[width=\linewidth,trim=2cm 1cm 2cm 3cm,clip]{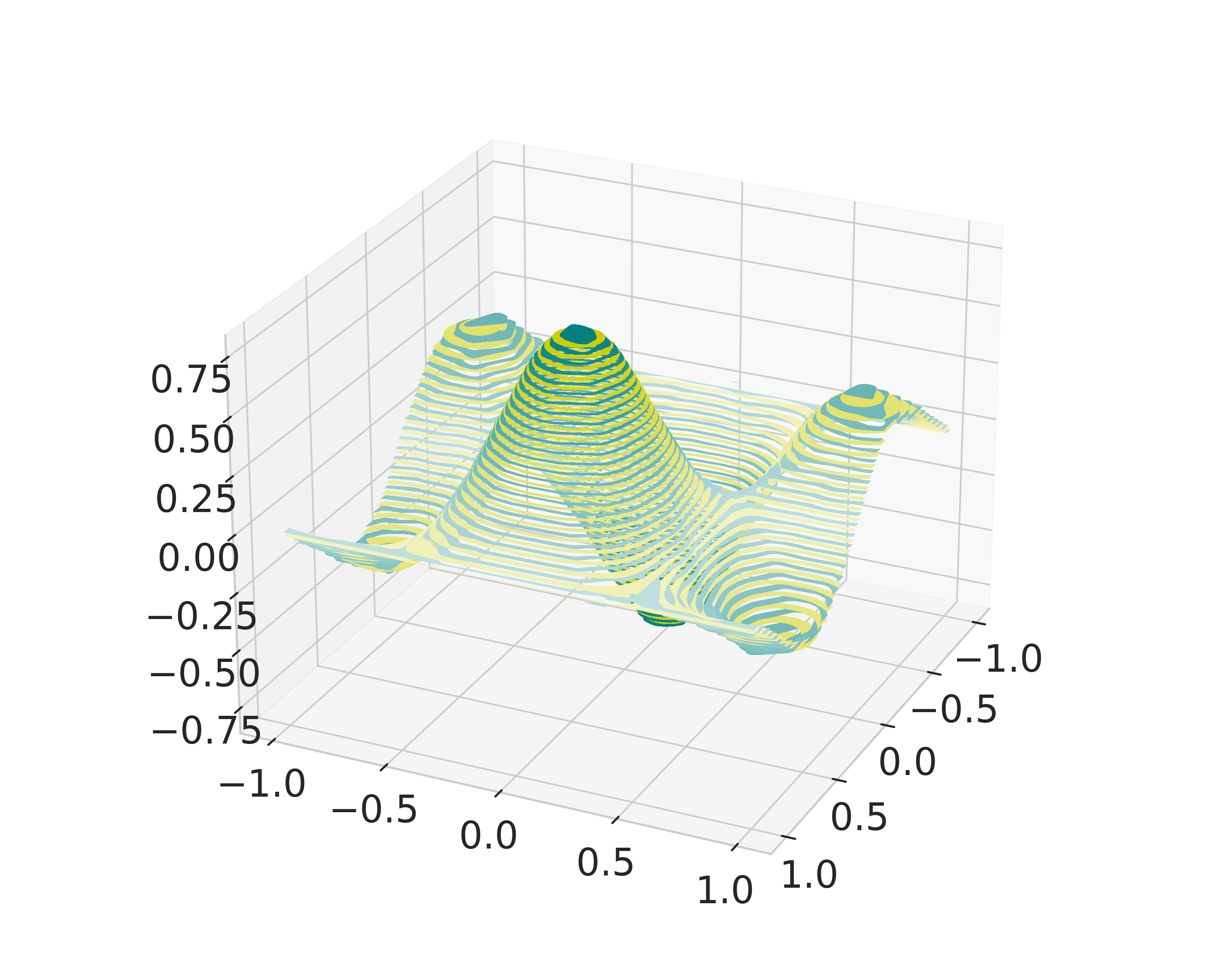}
		\caption{4.25e-02}
		\label{fig:ex3_appr_ga_init}
	\end{subfigure}
	\begin{subfigure}{.32\linewidth}
		\includegraphics[width=\linewidth,trim=2cm 1cm 2cm 3cm,clip]{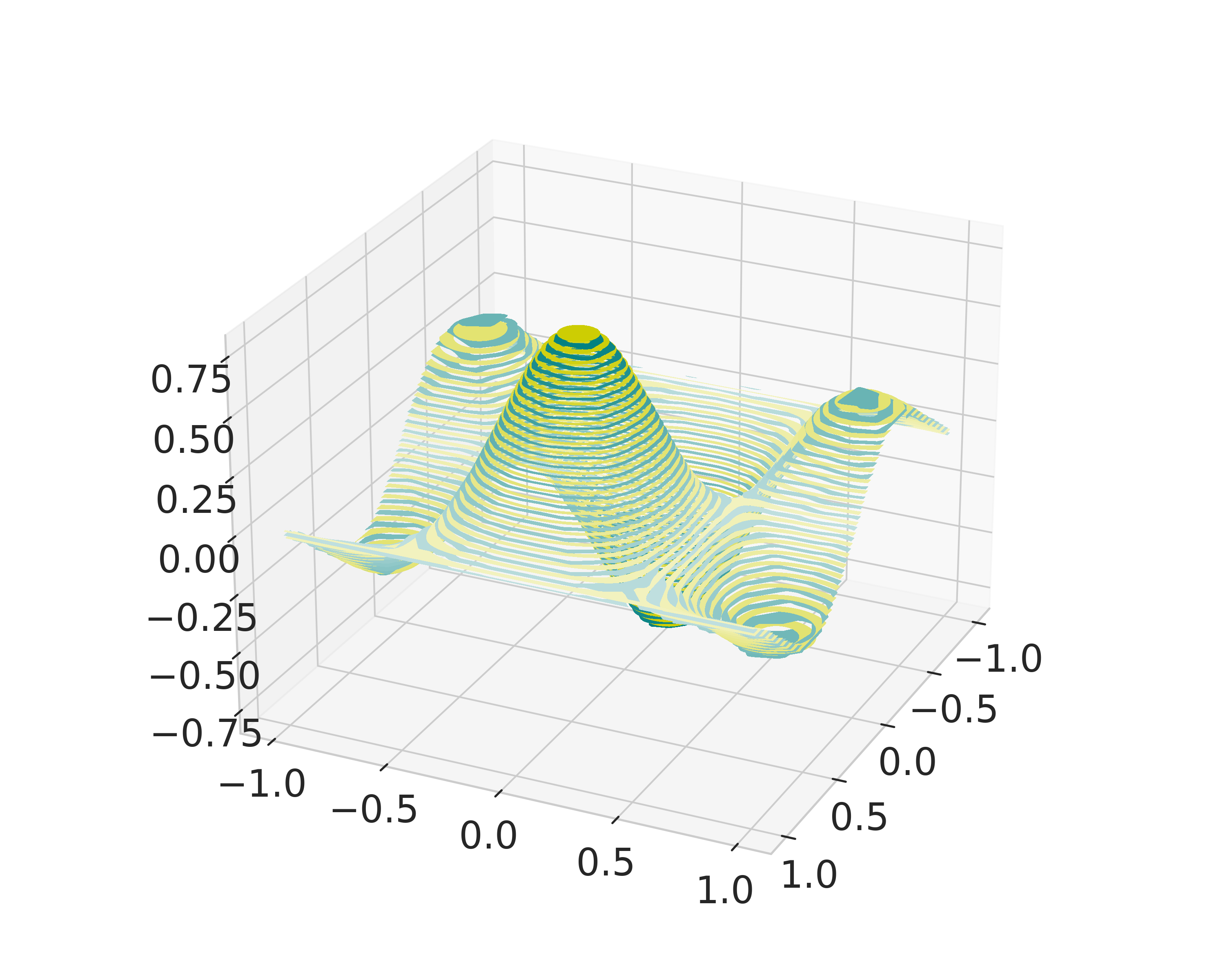}
		\caption{3.31e-02}
		\label{fig:ex3_appr_ga_train}
	\end{subfigure}
	\begin{subfigure}{.32\linewidth}
		\includegraphics[width=\linewidth,trim=2cm 1cm 2cm 3cm,clip]{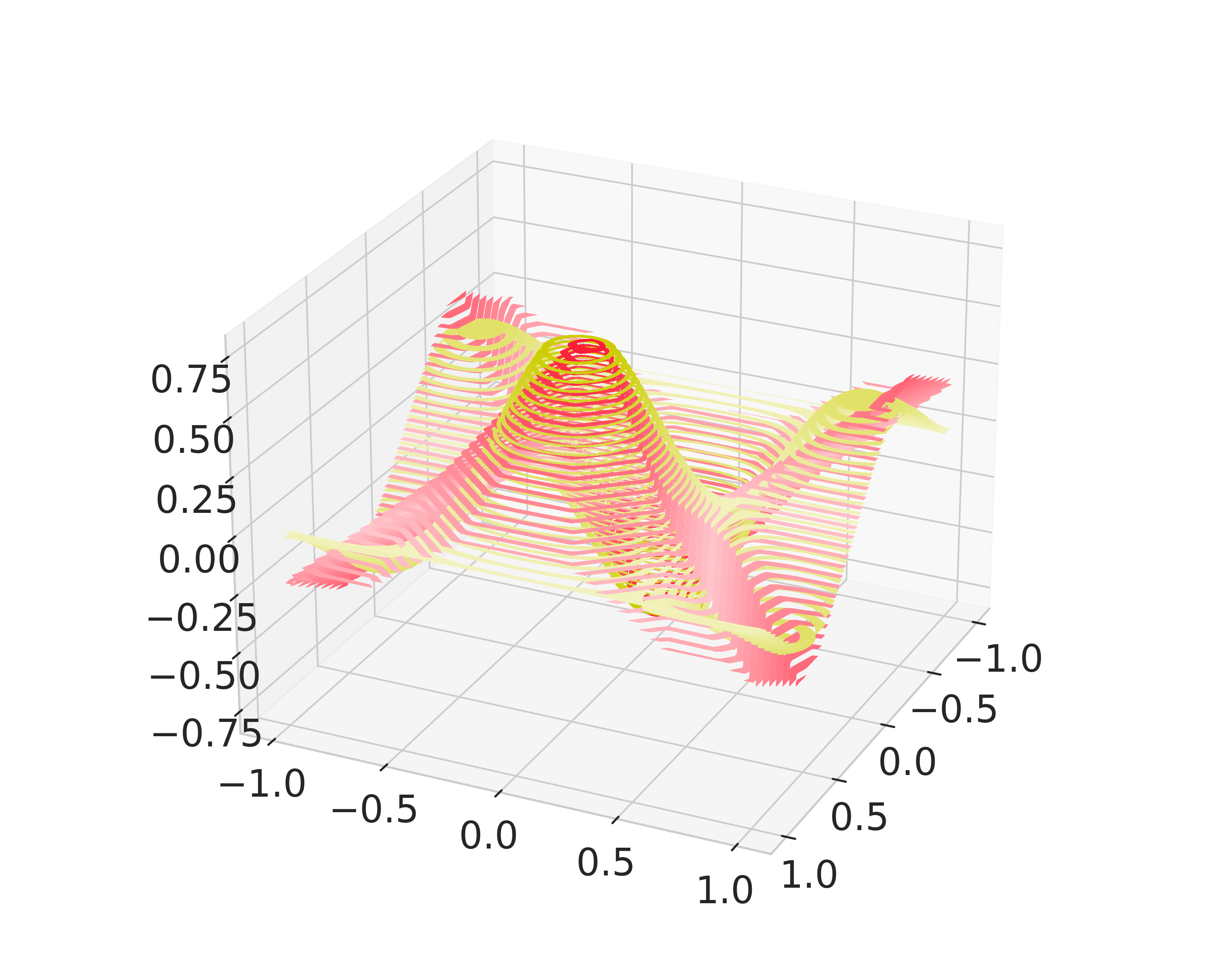}
		\caption{2.25e-01}
		\label{fig:ex3_appr_rnd}
	\end{subfigure}
	\caption{(a) GSN initialization, (b) network trained with GSN initialization, (c) network trained with random initialization. Values under the images indicate the $\ell_2$-approximation error on the test set.}
	\label{fig:ex3_appr}
\end{figure}

\begin{figure}[htbp]
	\centering
	\begin{subfigure}{.32\linewidth}
		\includegraphics[width=\linewidth,trim=3cm 2cm 2cm 3cm,clip]{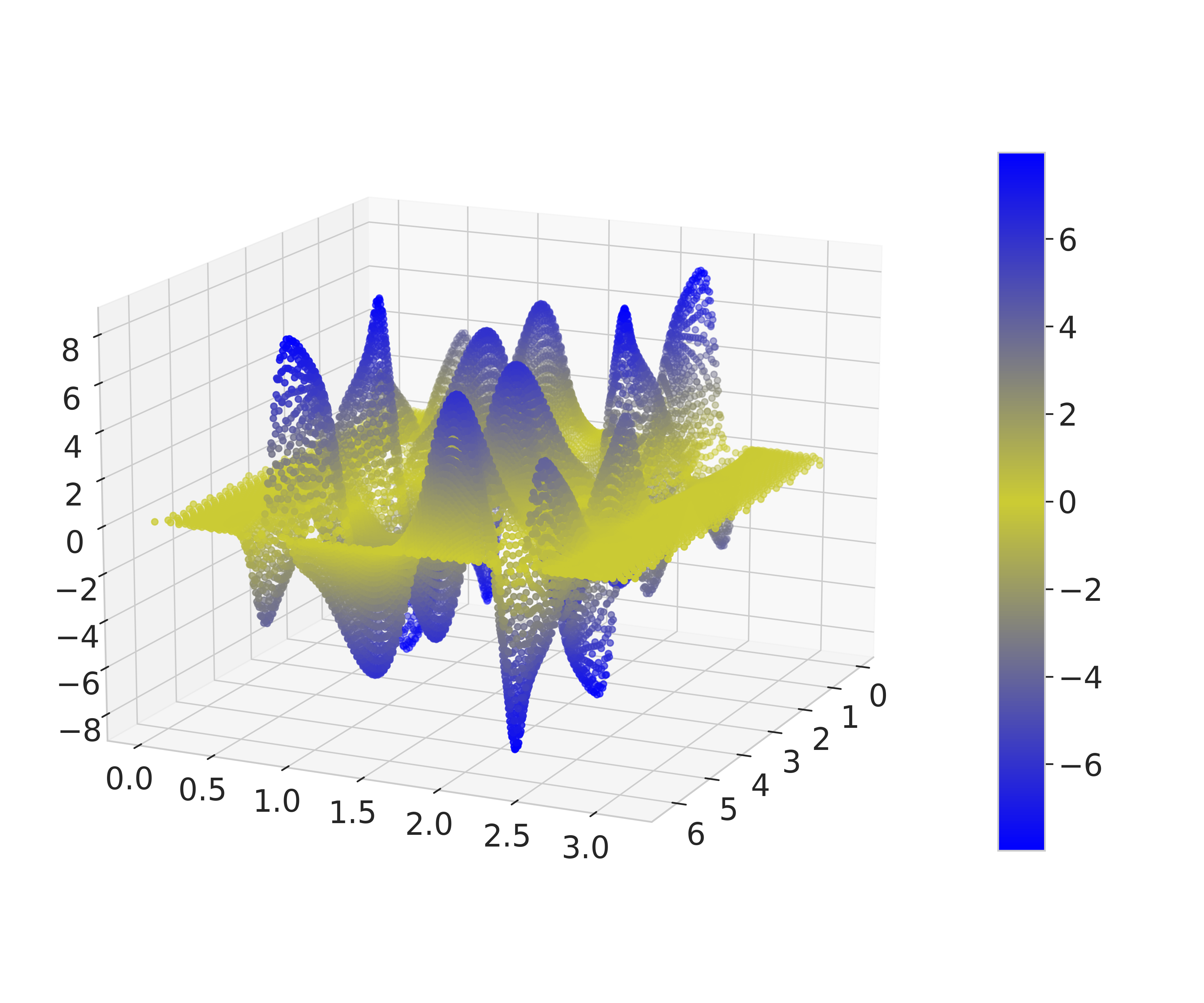}
		\caption{$\mathcal{CR}f(\phi)$}
		\label{fig:ex3_crf}
	\end{subfigure}
	\begin{subfigure}{.32\linewidth}
		\includegraphics[width=\linewidth]{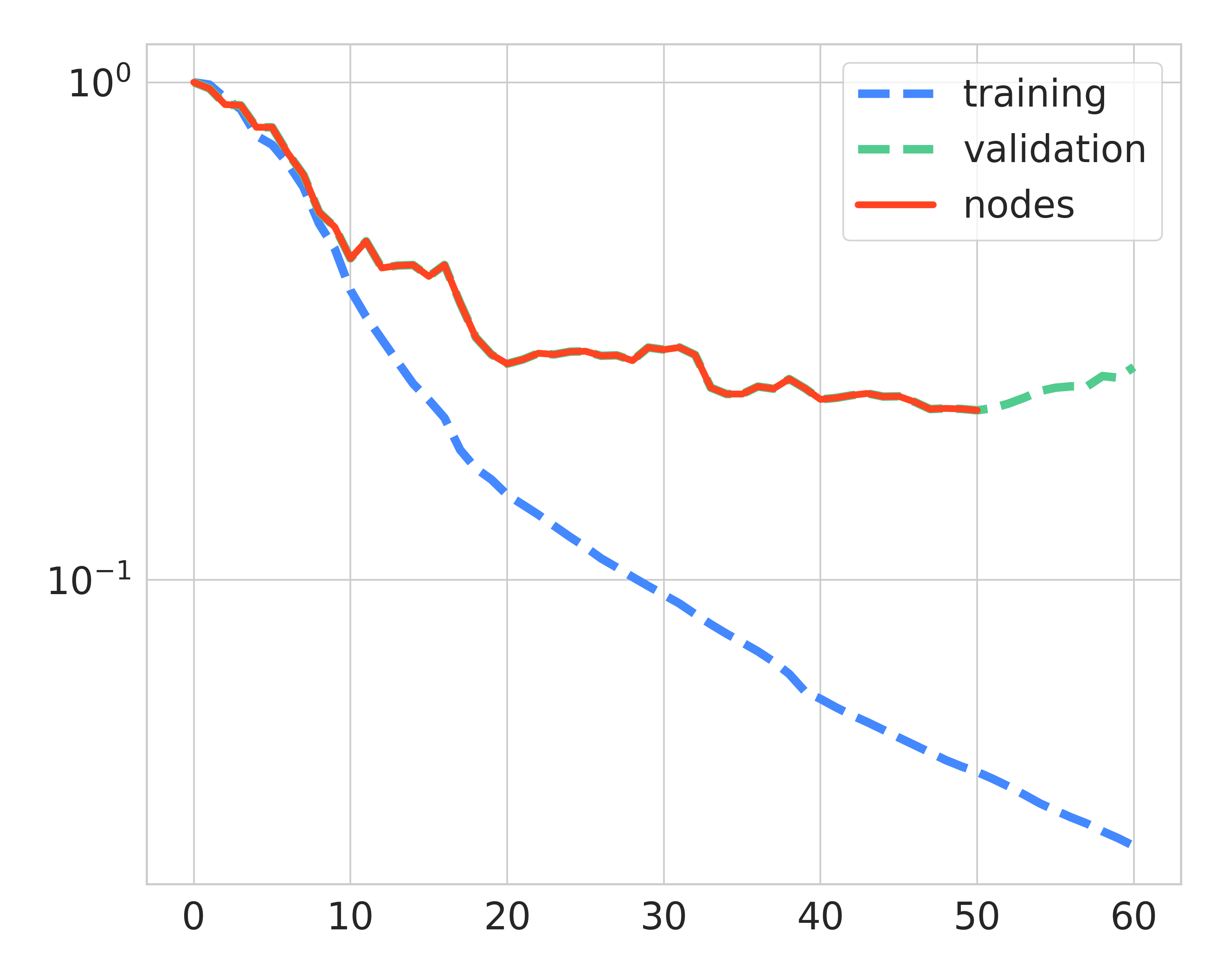}
		\caption{\#nodes: $50$}
		\label{fig:ex3_oga}
	\end{subfigure}
	\begin{subfigure}{.32\linewidth}
		\includegraphics[width=\linewidth]{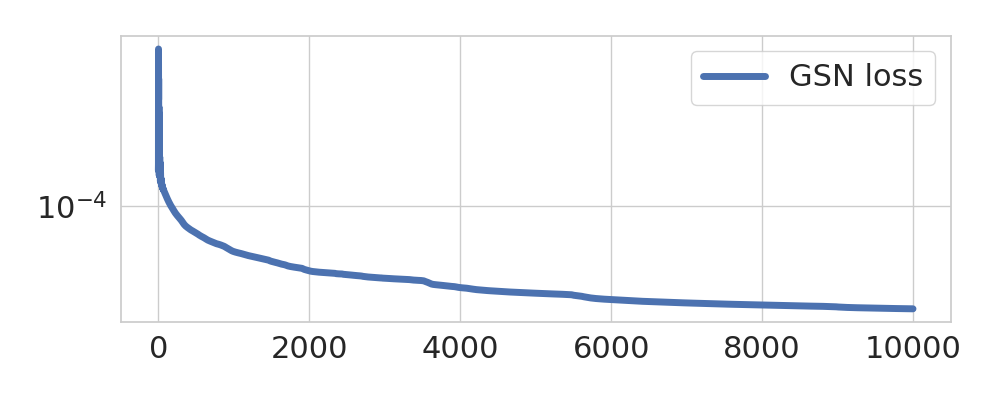}
		\includegraphics[width=\linewidth]{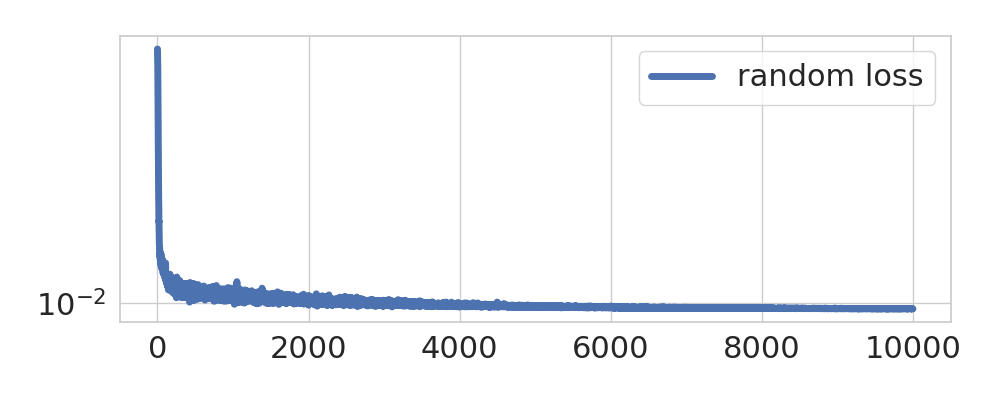}
		\caption{training losses}
		\label{fig:ex3_loss}
	\end{subfigure}
	\caption{Construction of the Shallow Greedy Network: (a) collapsed ridgelet transform of $f$, (b) nodes selection by greedy algorithm, (c) training of the neural network with GSN initialization (top) / random initialization (bottom).}
\end{figure}

Consider the target function $f : [-1,1]^2 \to \mathbb{R}$ given by
\[
	f(x,y) = \sin(\pi x) \cos(\pi y) \, \exp(-(x^2 + y^2)).
\]
We take $256$ training points, $50$ validation points, and $10,000$ test points.
The batch size is set to be $256$ for the network with the GSN initialization and $3$ for the network with random initialization.
The resulting approximations are presented in Figure~\ref{fig:ex3_appr}.
The collapsed ridgelet transform is shown in Figure~\ref{fig:ex3_crf}, the number of constructed nodes is $50$ (see Figure~\ref{fig:ex3_oga}).

\subsection{Example 4}\label{sec:ex4}
\begin{figure}[htbp]
    \vspace*{-.1in}
	\centering
	\begin{subfigure}{.32\linewidth}
		\includegraphics[width=\linewidth,trim=2cm 1cm 2cm 3cm,clip]{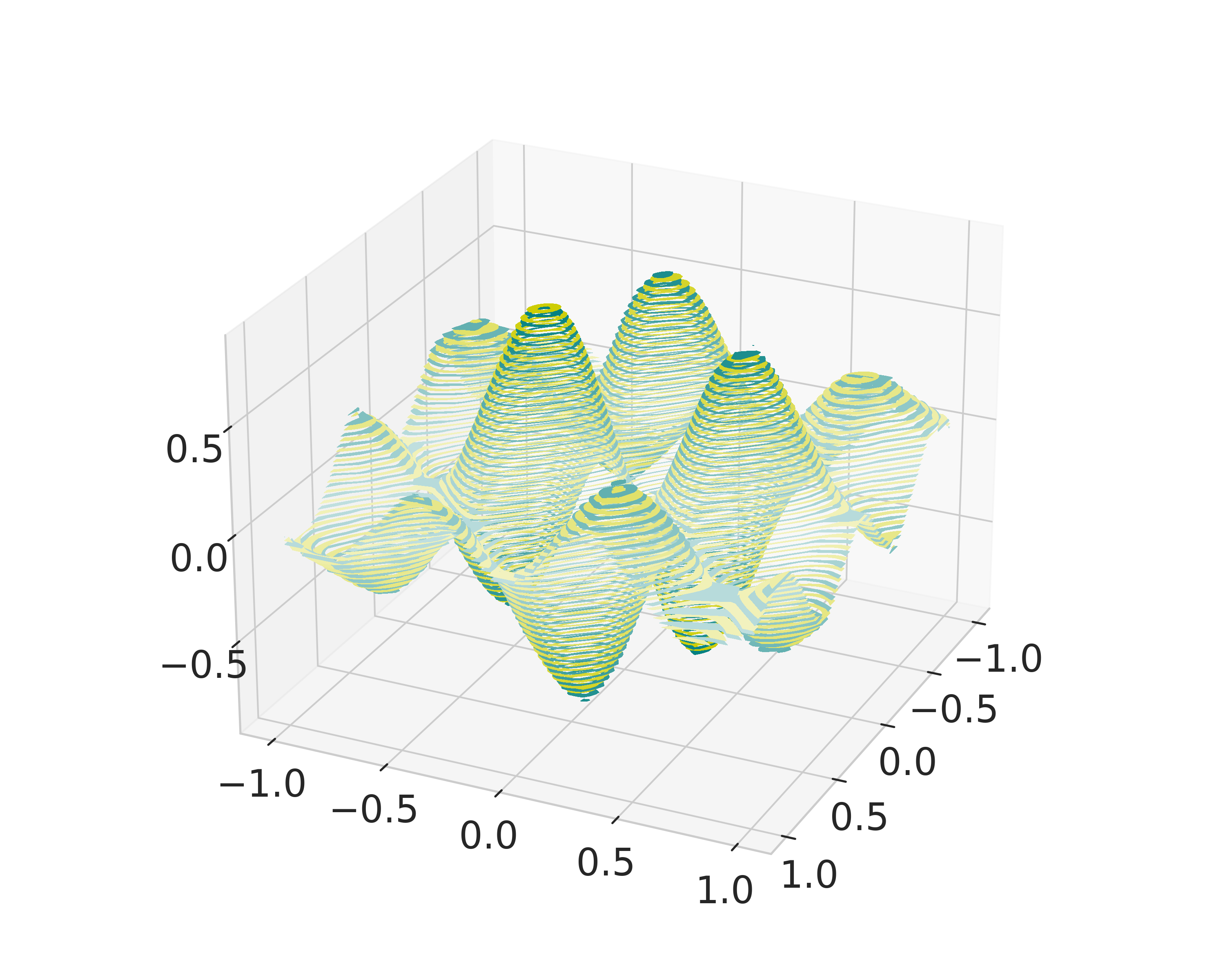}
		\caption{4.26e-02}
		\label{fig:ex4_appr_ga_init}
	\end{subfigure}
	\begin{subfigure}{.32\linewidth}
		\includegraphics[width=\linewidth,trim=2cm 1cm 2cm 3cm,clip]{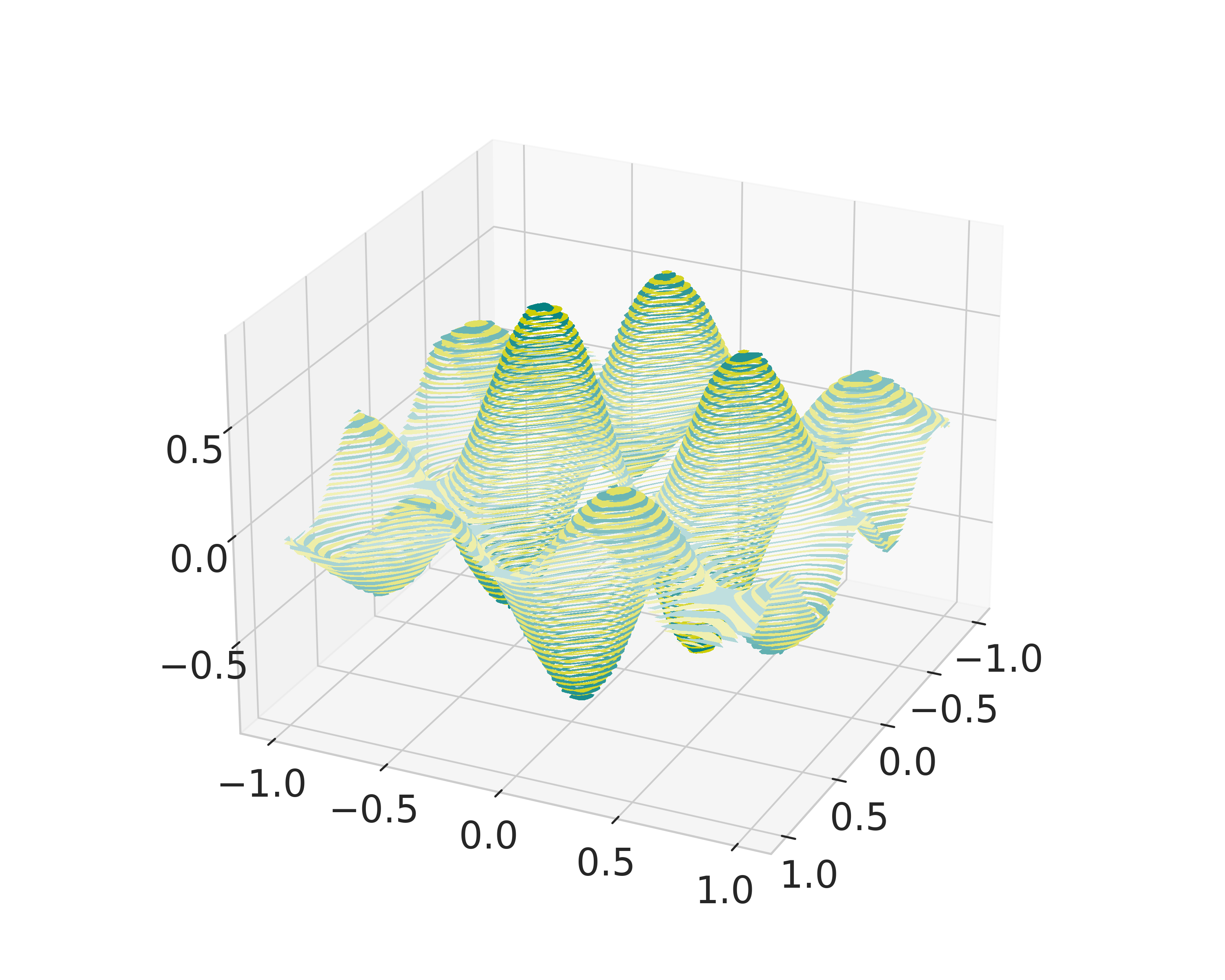}
		\caption{3.59e-02}
		\label{fig:ex4_appr_ga_train}
	\end{subfigure}
	\begin{subfigure}{.32\linewidth}
		\includegraphics[width=\linewidth,trim=2cm 1cm 2cm 3cm,clip]{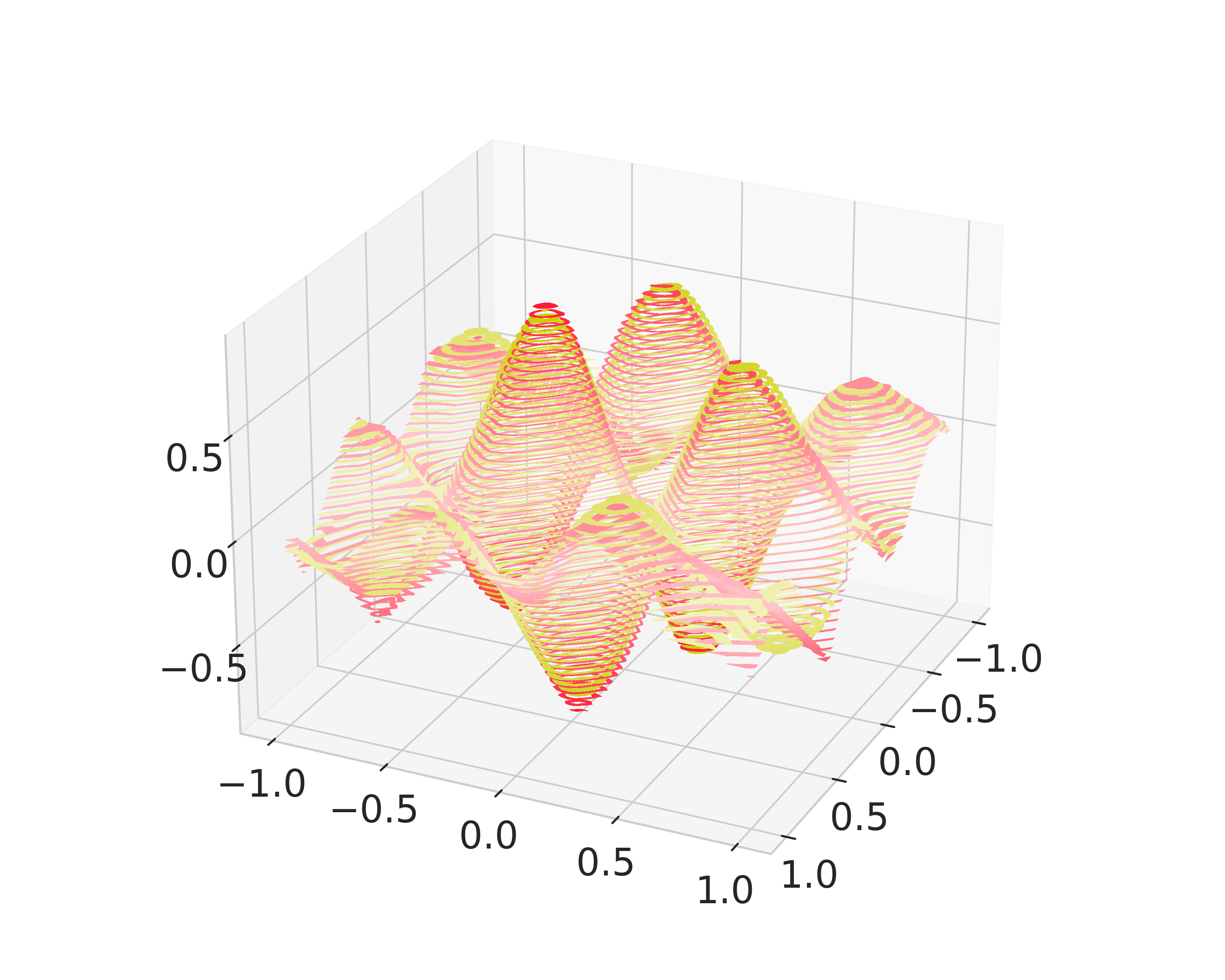}
		\caption{2.14e-01}
		\label{fig:ex4_appr_rnd}
	\end{subfigure}
	\caption{(a) GSN initialization, (b) network trained with GSN initialization, (c) network trained with random initialization. Values under the images indicate the $\ell_2$-approximation error on the test set.}
	\label{fig:ex4_appr}
\end{figure}

\begin{figure}[htbp]
	\centering
	\begin{subfigure}{.32\linewidth}
		\includegraphics[width=\linewidth,trim=3cm 2cm 2cm 3cm,clip]{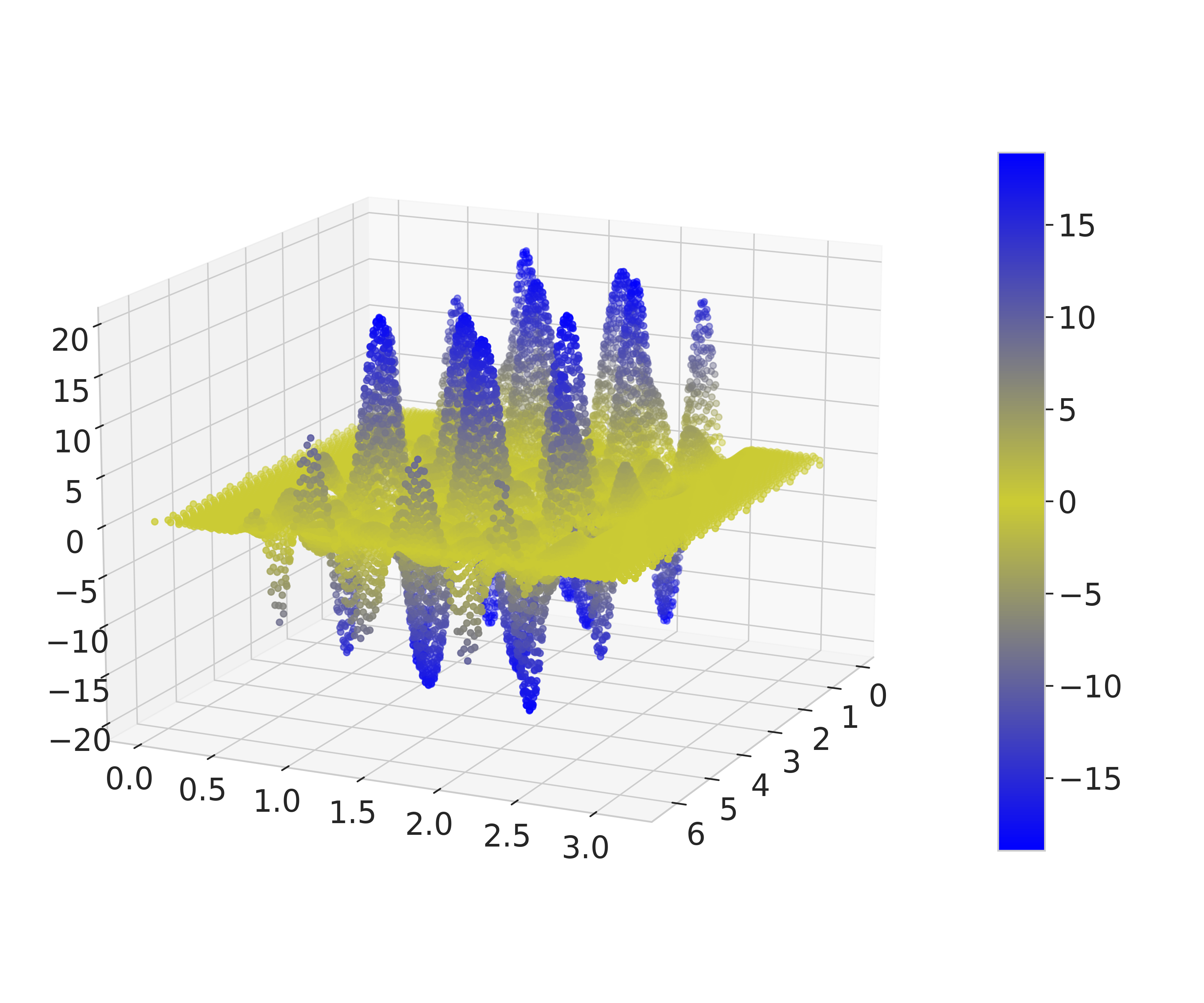}
		\caption{$\mathcal{CR}f(\phi)$}
		\label{fig:ex4_crf}
	\end{subfigure}
	\begin{subfigure}{.32\linewidth}
		\includegraphics[width=\linewidth]{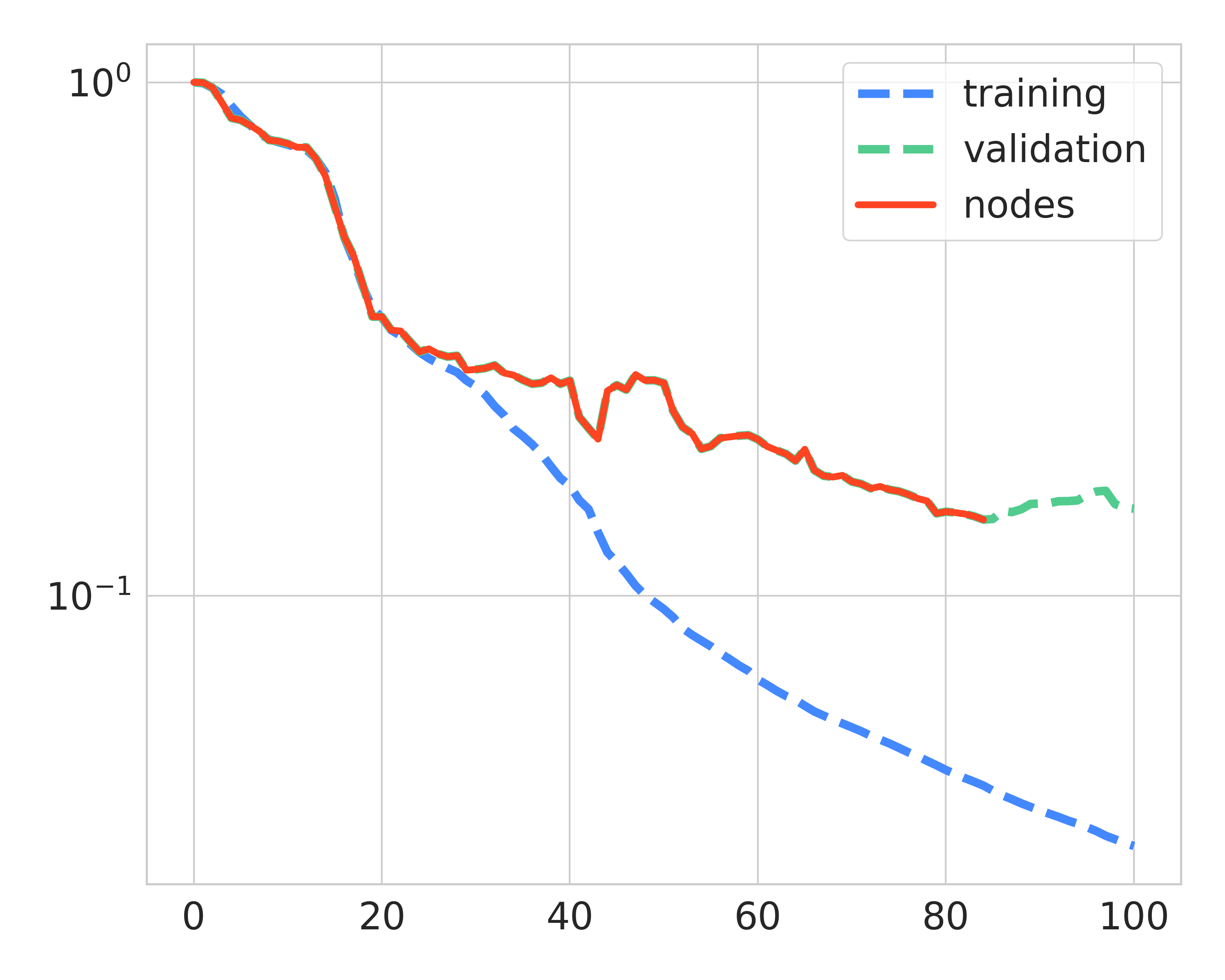}
		\caption{\#nodes: $84$}
		\label{fig:ex4_oga}
	\end{subfigure}
	\begin{subfigure}{.32\linewidth}
		\includegraphics[width=\linewidth]{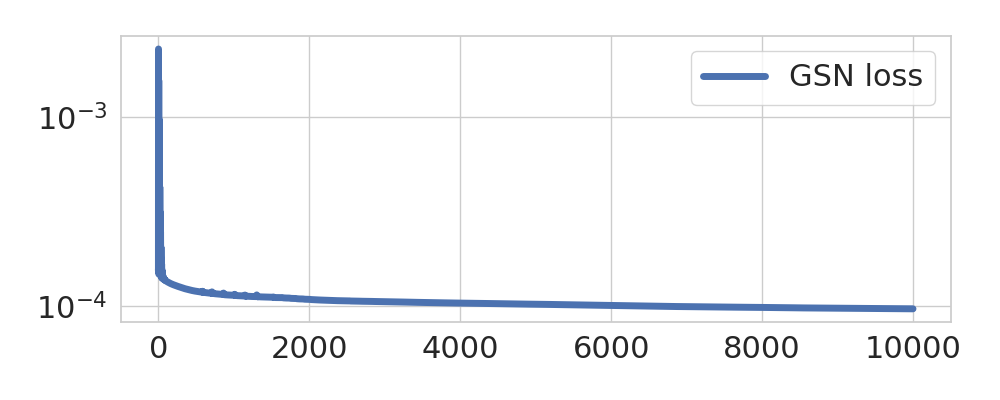}
		\includegraphics[width=\linewidth]{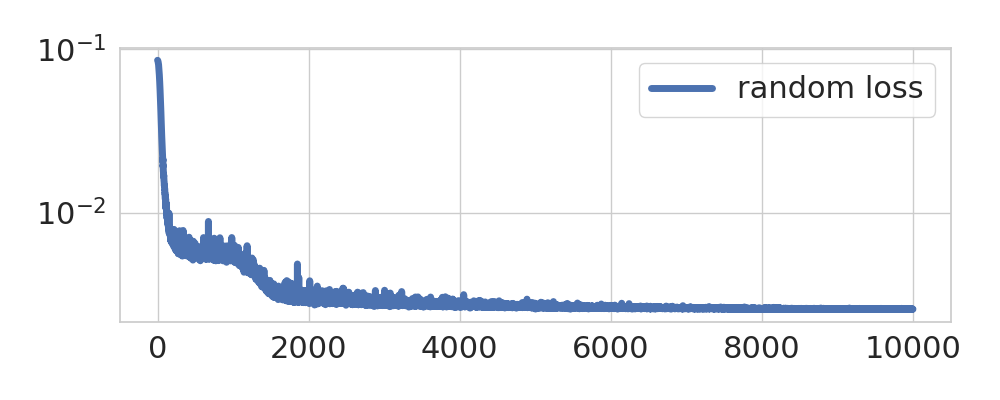}
		\caption{training losses}
		\label{fig:ex4_loss}
	\end{subfigure}
	\caption{Construction of the Shallow Greedy Network: (a) collapsed ridgelet transform of $f$, (b) nodes selection by greedy algorithm, (c) training of the neural network with GSN initialization (top) / random initialization (bottom).}
\end{figure}

Consider the target function $f : [-1,1]^2 \to \mathbb{R}$ given by
\[
	f(x,y) = \cos(5(x+y)) \sin(3(x-y)) \, \exp(-(x^2 + y^2)).
\]
We take $1,024$ training points, $200$ validation points, and $10,000$ test points.
The batch size is set to be $1,024$ for the network with the GSN initialization and $11$ for the network with random initialization.
The resulting approximations are presented in Figure~\ref{fig:ex4_appr}.
The collapsed ridgelet transform is shown in Figure~\ref{fig:ex4_crf}, the number of constructed nodes is $84$ (see Figure~\ref{fig:ex4_oga}).

\subsection{Example 5}\label{sec:ex5}
\begin{figure}[htbp]
	\centering
	\begin{subfigure}{.32\linewidth}
		\includegraphics[width=\linewidth]{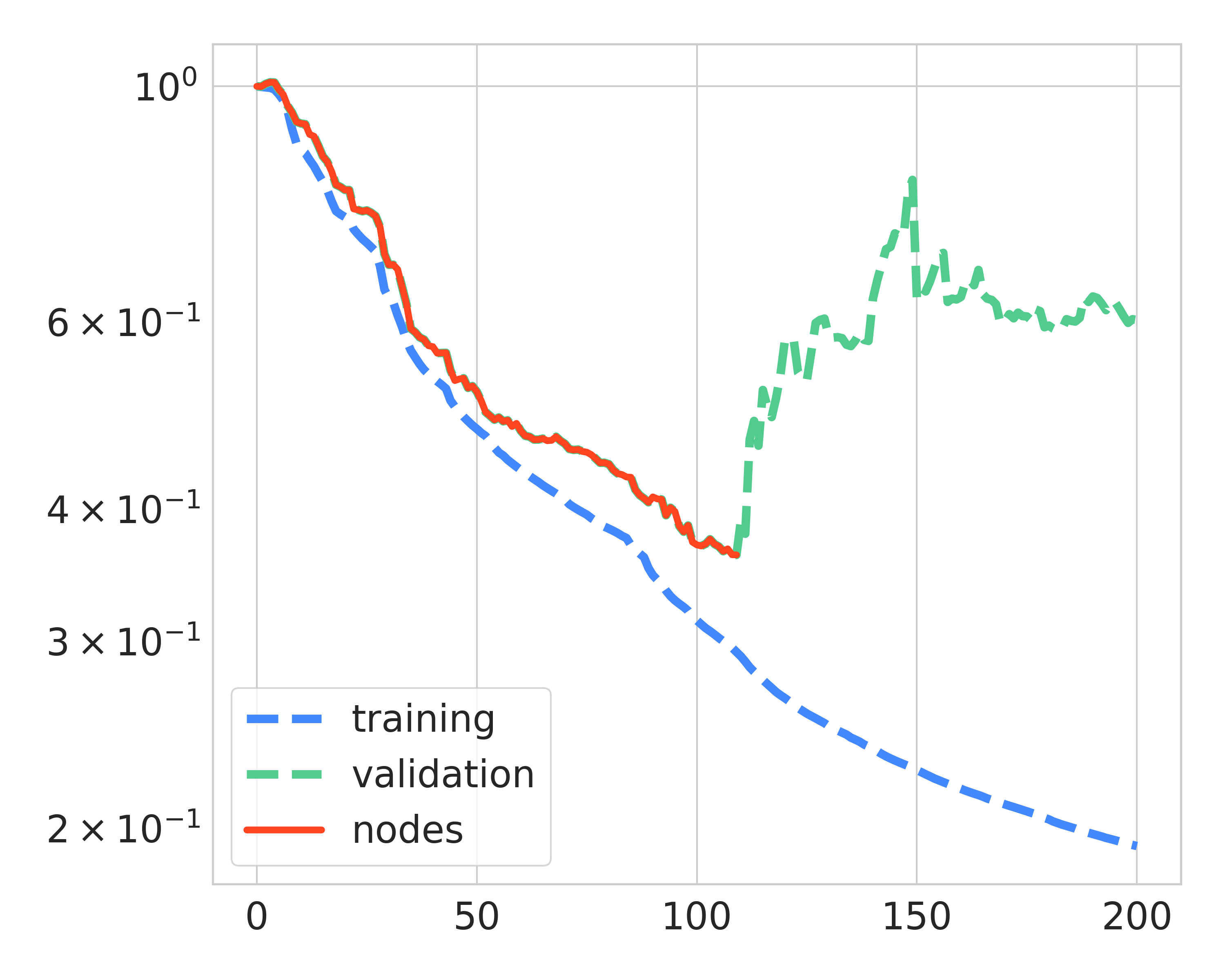}
		\caption{\#nodes: $109$}
		\label{fig:ex5_oga}
	\end{subfigure}
	\hspace*{.4in}
	\begin{subfigure}{.32\linewidth}
		\includegraphics[width=\linewidth]{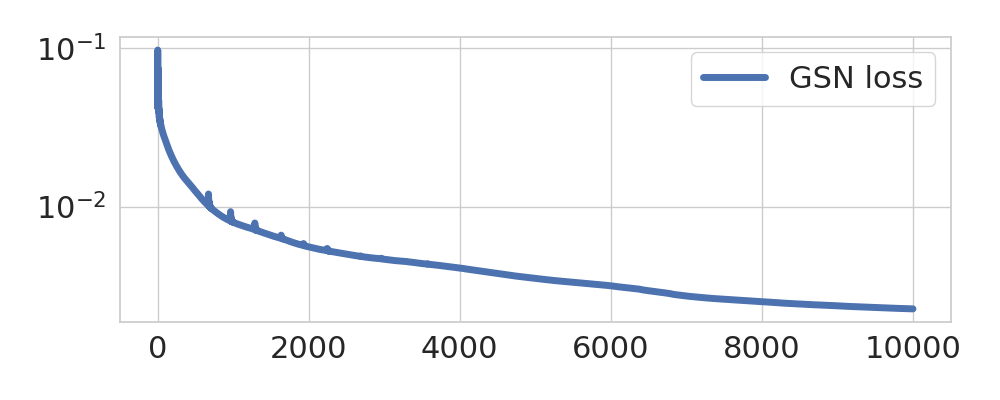}
		\includegraphics[width=\linewidth]{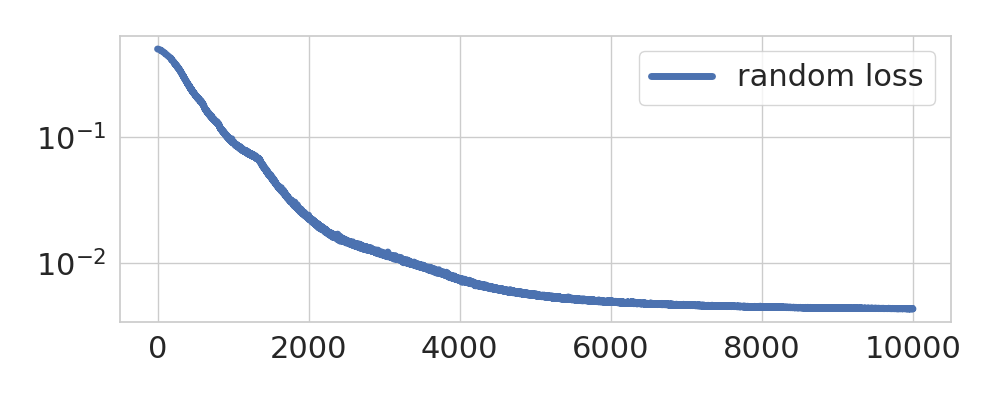}
		\caption{training losses}
		\label{fig:ex5_loss}
	\end{subfigure}
	\caption{(a) nodes selection by greedy algorithm, (b) training of the neural network with GSN initialization (top) / random initialization (bottom).}
\end{figure}

Consider the target function $f : [-1,1]^4 \to \mathbb{R}$ given by
\[
	f(\bar{x}) = \sin(2\pi(x_1 + x_2 + x_3 + x_4)).
\]
We take $4,000$ training points, $400$ validation points, and $10,000$ test points.
The number of constructed nodes is $109$ (see Figure~\ref{fig:ex5_oga}).

The approximation errors on the test set are the following: 3.20e-01 (GSN initialization), 6.99e-02 (network trained with the GSN initialization), 4.29e-01 (network trained with random initialization).
Note that in this example the GSN initialization by itself provides lower accuracy than the network trained with random initialization, which is caused by a higher dimensionality of the training data and the iterative nature of the greedy algorithm.
The network trained with the GSN initialization, however, still achieves significantly better accuracy when compared to a random initialization.

\section{Discussion}
In this paper we propose a novel way of initializing shallow neural networks with the ReLU activation via an iterative greedy strategy.
GSN initialization provides an interpretable way of selecting an architecture by allowing for a measurable trade-off between the number of nodes and the approximation accuracy.
Due to the nature of greedy selection, the GSN is computationally efficient in the sense that each constructed node is essential and contributes to the approximation of a target function.

Presented numerical results show that initializing a network with our approach stabilizes and improves the training of a shallow network.
In particular, greedy initialization allows us to use the full training set as one batch and run the optimization algorithm for a much fewer number of epochs, which we did not find possible with the randomly initialized networks.
Moreover, since, for a chosen set of parameters, the GSN initialization is deterministic, our approach is not affected by the issue of the initialization sensitivity.
Additionally, we observe that in some cases the GSN initialization can serve as a fully-trained neural network, thus potentially removing the need of learning via the backpropagation.

We also note that the approximation accuracy in the presented numerical examples is by no means the best our method can achieve; the purpose of the numerical section is to compare the GSN initialization with the classical approach in terms of training stability and hyperparameter dependency. 

While in this paper we only consider the ReLU activation, our method can be employed with any other positively homogeneous activation function.
For a wider range of activations, the appropriate modifications are required, which is one of the topics of our future research projects.
Similarly, the currently employed $\ell_2$-norm can be replaced with any other norm, which is in immediate interest for various applications.
Finally, we mention that even though the current method cannot be directly extended for the case of deep neural networks, the work in this direction is our main priority.


\section*{Acknowledgment}
This material is based upon work supported in part by: the U.S. Department of Energy, Office of Science, Early Career Research Program under award number ERKJ314; U.S. Department of Energy, Office of Advanced Scientific Computing Research under award numbers ERKJ331 and ERKJ345; the National Science Foundation, Division of Mathematical Sciences, Computational Mathematics program under contract number DMS1620280; and by the Laboratory Directed Research and Development program at the Oak Ridge National Laboratory, which is operated by UT-Battelle, LLC., for the U.S. Department of Energy under contract DE-AC05-00OR22725.


{\small
\bibliographystyle{abbrv}
\bibliography{biblio}
}


\end{document}